\documentclass[11pt]{article}
\PassOptionsToPackage{table}{xcolor}

\usepackage[final]{acl}
\usepackage{times}
\usepackage{latexsym}
\usepackage[T1]{fontenc}
\usepackage[utf8]{inputenc}
\usepackage{amsmath}
\usepackage{amssymb}
\usepackage{microtype}
\usepackage{inconsolata}
\usepackage{graphicx}
\usepackage{booktabs}
\usepackage{enumitem}
\usepackage{multirow}
\usepackage{algorithm}
\usepackage{algorithmic}
\usepackage{listings}
\usepackage{tcolorbox}
\tcbuselibrary{skins, listings, breakable}
\allowdisplaybreaks
\makeatletter
\def\@makefnmark{\hbox{}}
\makeatother

\title{FormuEvo: LLM-Guided Evolution for Discovering Solver-Efficient Mixed-Integer Programming Formulations}

\author{
 \textbf{Haofeng Yuan\textsuperscript{1}},
 \textbf{Jianing Peng\textsuperscript{1}},
 \textbf{Jieyi Bi\textsuperscript{2}},
 \textbf{Ni Zhang\textsuperscript{3}},
 \textbf{Shiji Song\textsuperscript{1,\,\dag}},
 \textbf{Zhiguang Cao\textsuperscript{3}}
\\
\\[-2pt]
\textsuperscript{1}Department of Automation, BNRist, Tsinghua University\\
\textsuperscript{2}College of Computing and Data Science, Nanyang Technological University\\
\textsuperscript{3}School of Computing and Information Systems, Singapore Management University
\\[2pt]
{\small
\texttt{yhf22@mails.tsinghua.edu.cn},\,
\texttt{pjn24@mails.tsinghua.edu.cn},\,
\texttt{jieyi001@e.ntu.edu.sg}
}\\[-2pt]
{\small
\texttt{ni.zhang.2025@phdcs.smu.edu.sg},\,
\texttt{shijis@mail.tsinghua.edu.cn},\,
\texttt{zgcao@smu.edu.sg}
}
}

\begin{document}
\maketitle
\footnotetext{\hspace*{-1.8em}\textsuperscript{\dag}Corresponding author.}
\footnote{\hspace*{-1.8em}\hphantom{\textsuperscript{\dag}}Code: \url{https://github.com/Xyz-yuanhf/formuevo}}

\begin{abstract}
Mixed-integer programming (MIP) lies at the core of operations research and industrial optimization. While large language models (LLMs) have recently shown promise in automated MIP modeling from natural language, they prioritize semantic correctness but overlook formulation strength, severely bottlenecking the efficiency of downstream solvers. We propose FormuEvo, an LLM-guided evolutionary framework for automated discovery of solver-efficient MIP formulations. FormuEvo frames MIP formulation design as evolutionary optimization over the symbolic space of MIP formulations, represented as executable modeling programs, by iteratively generating, evaluating, and selecting stronger candidates via LLM-driven crossover, mutation, and repair operations. To move beyond blind exploration, FormuEvo introduces a solver-informed diagnosis mechanism that exploits fine-grained solver statistics as verbal gradients for targeted refinement. Additionally, a structured memory abstracts prior experience into reusable modeling strategies, avoiding redundant exploration while enabling zero-shot transfer to unseen problems and bootstrapping smaller LLMs. Experiments across diverse linear and non-linear problems demonstrate that FormuEvo discovers formulations that significantly outperform both expert-designed formulations and existing LLM-based approaches, accelerating solvers by up to $5.5\times$, with distilled knowledge transferring effectively across problems and model scales.
\end{abstract}

\section{Introduction}

\begin{figure}[!t]
\centering
\includegraphics[width=0.97\linewidth]{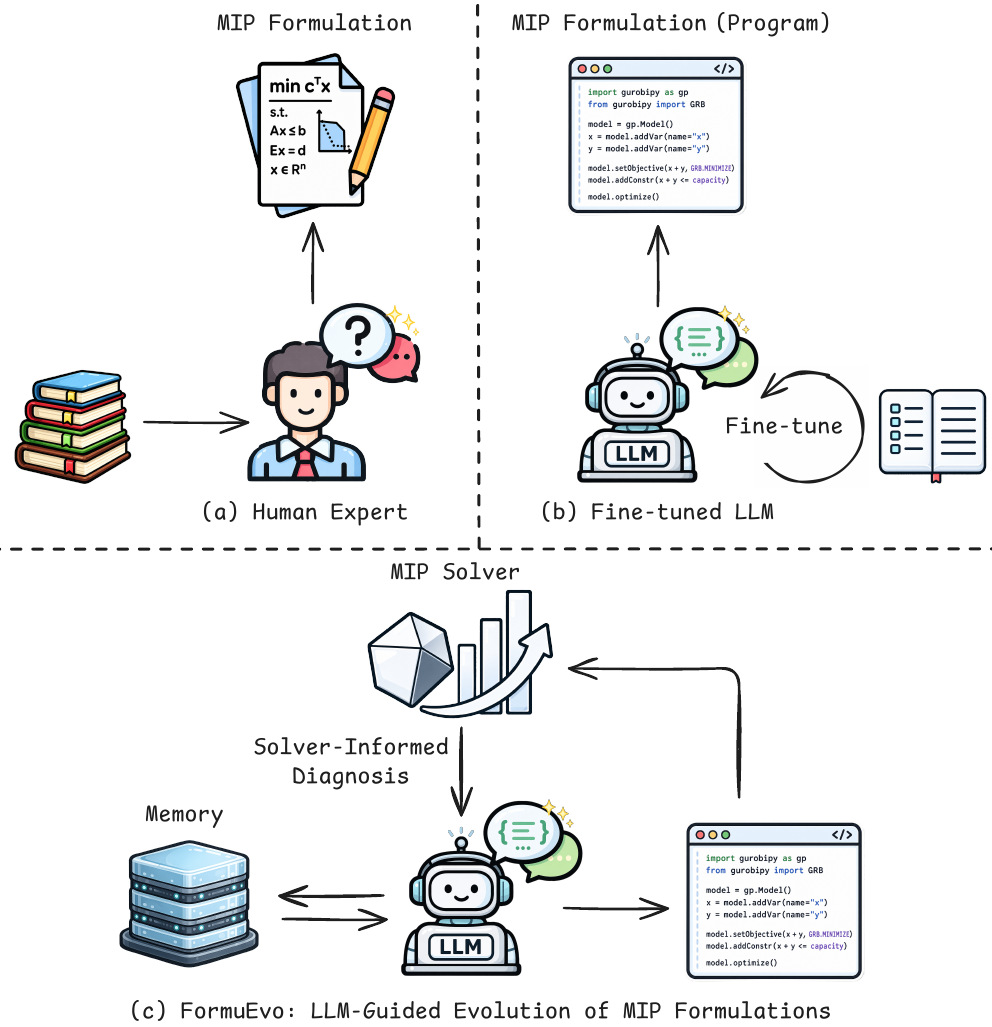}
\caption{MIP formulation design via (a) human experts, (b) LLM fine-tuning with end-to-end generation, and (c) the LLM-guided evolution framework of FormuEvo.}
\label{fig:overview}
\end{figure}

Mixed-integer programming (MIP) plays a fundamental role in decision-making across domains such as manufacturing, transportation, and management~\cite{bixby2012brief,wolsey2020integer,clautiaux2025last}. The practical success of MIP has been largely driven by decades of advances in powerful solvers such as Gurobi~\cite{gurobi2026gurobi} and COPT~\cite{ge2022cardinal}, which integrate sophisticated algorithmic components including branch-and-bound, cutting planes, and various heuristics to solve problems of growing scale and complexity.

A well-established principle in mathematical programming is that a single optimization problem may admit multiple mathematically equivalent MIP formulations, which guarantee the same optimal solution but can differ by orders of magnitude in computational efficiency~\cite{wolsey2020integer}. Consequently, the practical tractability of MIP depends not only on constructing a mathematically \textit{correct} formulation, but also on designing one that is computationally \textit{strong}, namely, whose structural properties enable MIP solvers to operate efficiently.

Despite its fundamental importance, discovering solver-efficient MIP formulations has long been regarded as one of the most challenging and expertise-intensive problems in operations research~\cite{vielma2015mixed}. Even experienced experts require substantial domain knowledge and structural insights to craft a strong formulation (Fig.~\hyperref[fig:overview]{\ref*{fig:overview}(a)}). This challenge is further complicated by the fact that modeling intuitions can become outdated: with the development of modern MIP solvers, classical modeling strategies once regarded as ``best practices'', such as certain strengthening constraints or symmetry-breaking rules, can inadvertently interfere with the advanced presolving procedures, cut generation, and heuristic routines, leading to counter-intuitive performance degradation~\cite{achterberg2013mixed,salvagnin2018symmetry}. This motivates a new automated approach capable of discovering solver-efficient MIP formulations and adaptable to the advancing algorithmic internals of modern solvers.

Large language models (LLMs) have recently demonstrated remarkable promise in automating MIP modeling from natural language~\cite{wang2025ormind,qian2025modelingagent,zhou2026steporlm}. By fine-tuning based on curated formulation datasets, they have effectively bridged the gap between natural language problem descriptions and executable MIP modeling programs via end-to-end generation (Fig.~\hyperref[fig:overview]{\ref*{fig:overview}(b)}). However, their objective is \textit{correctness} and \textit{executability}, namely, producing a formulation that accurately encodes the target problem and can be successfully solved by a downstream solver. As a result, the formulations they generate, though correct, are often computationally naive, falling far short of strong formulations required for large-scale real-world optimization.

To address this gap, we propose \textbf{FormuEvo}, a novel LLM-guided evolutionary framework for automated discovery of solver-efficient MIP formulations. Rather than treating formulation design as single-pass generation, FormuEvo reframes it as \textit{evolutionary optimization} over the symbolic space of MIP formulations, which are represented as executable modeling programs. Given a natural language problem description and a formulation template, FormuEvo maintains a population of candidate formulations, iteratively applying LLM-driven crossover, mutation, and repair operations guided directly by the solver performance, progressively evolving toward solver-efficient formulations (Fig.~\hyperref[fig:overview]{\ref*{fig:overview}(c)}). To move beyond blind exploration, FormuEvo introduces two key mechanisms: 1) a \textit{solver-informed diagnosis} mechanism, where a diagnostic LLM converts fine-grained solver statistics into interpretable verbal gradients, guiding evolution toward targeted refinement, and 2) a \textit{structured memory}, which abstracts prior experience into reusable modeling strategies, avoiding redundant exploration while further enabling zero-shot transfer to unseen problems and bootstrapping smaller LLMs by distilling a generalizable knowledge base. Our main contributions are as follows:
\begin{itemize}[leftmargin=*, itemsep=1pt, topsep=3pt, parsep=0pt]
    \item We propose FormuEvo, an LLM-guided evolutionary framework for automated discovery of solver-efficient MIP formulations.
    \item We introduce a solver-informed diagnosis mechanism that leverages fine-grained solver statistics for targeted, solver-aware evolution.
    \item We develop a structured memory that distills reusable knowledge, enabling efficient search and transfer across problem and model scales.
    \item FormuEvo discovers non-obvious MIP formulations that outperform the state-of-the-art baselines, accelerating MIP solvers by up to $5.5\times$.
\end{itemize}

\section{Related Work}

\subsection{MIP Modeling and Formulation}
Designing strong MIP formulations is a longstanding challenge in operations research~\cite{wolsey2020integer}. Classical strengthening techniques include valid inequalities and cutting planes~\cite{nemhauser1988integer,cornuejols2008valid}, symmetry-breaking constraints~\cite{margot2009symmetry}, and extended variables and reformulations~\cite{conforti2013extended}. However, with rapid development of modern MIP solvers, classical techniques can sometimes conflict with sophisticated solver internals and lead to counter-intuitive performance degradation~\cite{achterberg2013mixed,salvagnin2018symmetry}, motivating the need for automated, adaptive approaches to formulation design.

\subsection{LLM for Automated MIP Modeling}
Recent advances in LLMs have automated MIP modeling from natural language through structured formulations~\cite{ramamonjison2022augmenting}, agent systems~\cite{ahmaditeshnizi2024optimus}, or fine-tuning~\cite{huang2025orlm,wang2026orprm,zhou2026steporlm}. Despite improving domain fidelity, they suffer from two major flaws. First, they operate at the \textit{instance} level, generating formulations with parameters tied to particular instances rather than discovering generalized, instance-agnostic formulations at the \textit{problem} level. Second, as these open-loop methods prioritize mathematical \textit{correctness} rather than formulation \textit{strength}, the generated formulations are structurally weak and inefficient for large-scale optimization.

\subsection{LLM-Guided Evolutionary Search}
LLM-guided evolutionary search has emerged as a powerful paradigm for automated algorithm design and heuristic discovery~\cite{liu2024evolution,novikov2025alphaevolve}. FunSearch~\cite{romera2024mathematical} pioneered this by using LLMs as evolutionary operators over executable code, while subsequent works~\cite{ye2024reevo,ye2025large,zhang2026an,hou2026llmbranch,gungordu2026pathwise} extended this framework for combinatorial heuristic design. However, little work has adapted this paradigm to MIP formulation. The most closely related work, EvoCut~\cite{yazdani2025evocut}, applies LLM-based evolution to cutting plane generation. However, EvoCut remains limited to local relaxation tightening on an existing model and lacks the capability of holistic structural redesign. In contrast, FormuEvo evolves over the global symbolic space of MIP formulations, and incorporates solver-informed diagnosis and structured memory to enable systematic and directed improvement.

\section{FormuEvo}

\subsection{Problem Formalization}

Given a problem $p$ described in natural language, an MIP formulation can be modeled as:
\begin{equation}
\begin{aligned}
\min_{x, y} \quad & c(x, y) \\
\mathrm{s.t.} \quad & g_i(x, y) \le 0, \quad i = 1, \dots, m, \\
& h_j(x, y) = 0, \quad j = 1, \dots, n, \\
& x \in X \subseteq \mathbb{R}^q, \\
& y \in Y \subseteq \mathbb{Z}^r,
\end{aligned}
\end{equation}
where $x$ and $y$ denote continuous and integer decision variables, $c(x, y)$ is the objective function, and $g_i(x, y)$, $h_j(x, y)$ denote inequality and equality constraints, respectively. In practice, the MIP formulation is instantiated as executable programs (e.g., Python code with Gurobi) that specify these components and can be directly executed by an MIP solver. We denote by $\mathcal{F}$ the symbolic space of MIP formulations of problem $p$, namely, the set of all semantically correct and syntactically valid programs encoding problem $p$. By definition, any two formulations $f_1,\, f_2 \in \mathcal{F}$, possibly involving different structures of variables and constraints, are equivalent in the sense of optimization, which guarantees the same optimal solution for problem $p$.
 
Existing LLM-based approaches treat MIP modeling as a \textit{satisfaction problem}, aiming to generate \textit{any} formulation that can be successfully solved by a downstream MIP solver and produce the correct solution. However, they largely overlook a fundamental principle in mathematical optimization: although formulations in $\mathcal{F}$ are mathematically equivalent, they can induce dramatically different solver behaviors and computational efficiency. In practice, two equivalent formulations may differ by orders of magnitude in runtime due to differences in relaxation tightness, symmetry structure, and even solver-specific implementation details. Motivated by this, we reframe MIP formulation design as an \textit{optimization problem} over the symbolic space of MIP formulations toward minimizing the downstream computational cost:
\begin{equation}
f^\star = \underset{f \in \mathcal{F}}{\arg\min}\; \phi(f),
\end{equation}
where $\phi(f)$ evaluates the computational cost (e.g., runtime) of formulation $f$ with a downstream MIP solver. This defines an optimization problem over a vast, discrete, and highly structured symbolic space, where the objective is non-differentiable and expensive to evaluate. Therefore, we frame the discovery of solver-efficient MIP formulations as an evolutionary search, with LLMs as operators to navigate the symbolic space of MIP formulations.

\subsection{LLM-Guided Evolutionary Search}

\begin{figure*}[!t]
\centering
\includegraphics[width=0.96\linewidth]{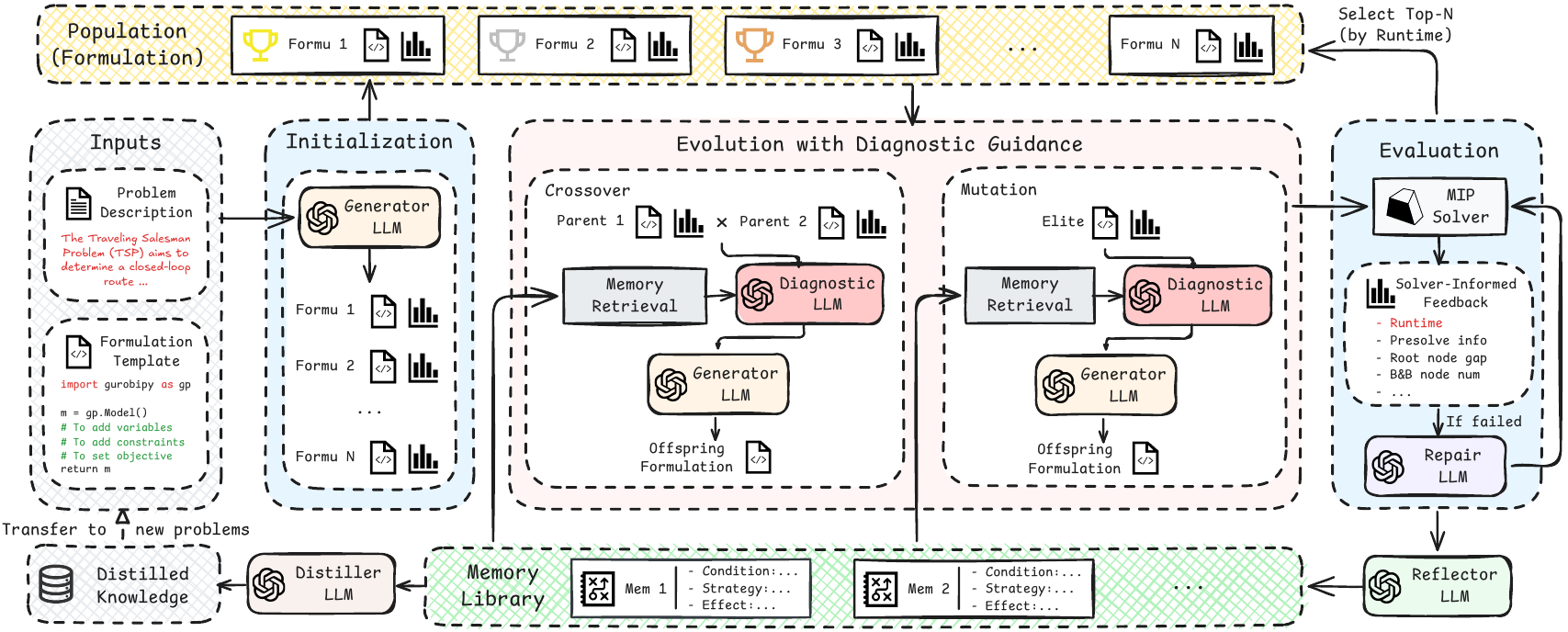}
\caption{The overview of FormuEvo. FormuEvo iteratively evolves formulations through LLM-driven crossover, mutation, and repair toward better solver efficiency, guided by solver-informed diagnosis and structured memory.}
\label{fig:framework}
\end{figure*}

FormuEvo takes a natural language description and a formulation template $f_0$ as input, maintaining a population $\mathcal{P}_t = \{f_1, \dots, f_N\}$ of $N$ candidate formulations at generation $t$. To effectively explore the vast and structured symbolic space $\mathcal{F}$, FormuEvo employs an evolutionary loop coordinated by five specialized LLM modules: 1) a \textit{generator LLM} that produces new candidate formulations, 2) a \textit{diagnostic LLM} that interprets solver feedback to guide generation, 3) a \textit{repair LLM} that debugs failed formulations, 4) a \textit{reflector LLM} that abstracts evolution experience into structural memories, and 5) a \textit{distiller LLM} that distills accumulated memory into transferable knowledge for unseen problems. The overall framework is illustrated in Fig.~\ref{fig:framework}.

\paragraph{Initialization.} Given the problem description and a formulation template, the evolution starts by prompting the generator LLM to produce $N$ candidate formulations as the initial population $\mathcal{P}_0$. The generator is encouraged to explore diverse formulation variants, such as alternative variables, logically equivalent but structurally different constraints, as well as variations in implementation details, to prevent premature convergence to sub-optimal local regions in the symbolic space.

\paragraph{Evaluation.} Each candidate formulation $f \in \mathcal{P}_t$ is evaluated by executing it on a downstream MIP solver (e.g., Gurobi) over a set of instances of problem $p$. If execution fails due to compilation errors or correctness violations (by verifying whether the solved objective values match the ground-truth optimal value), the error tracebacks and solver logs are fed into a repair LLM, which attempts to fix the candidate formulation while preserving its intended modeling logic. Candidates exceeding the maximum repair budget are discarded, ensuring that all maintained formulations strictly belong to $\mathcal{F}$. Notably, since the generator LLM may produce bold yet error-prone formulation variants, the repair operations can still be valuable for exploration.

For each successfully evaluated formulation, its fitness is measured by the solver cost $\phi(f)$, computed as the shifted geometric mean (SGM) of per-instance runtimes, a standard evaluation metric for solver performance. Meanwhile, the reflector LLM abstracts the modeling decisions and solver feedback from each evaluated formulation into reusable strategies, which are stored in a structured memory library for future retrieval (Section~\ref{sec:memory}).

\paragraph{Evolution.} At each generation, FormuEvo preserves the top-$N$ formulations according to their fitness. The next generation is produced through two LLM-based operators: \textit{crossover} and \textit{mutation}.

For crossover, two distinct parent formulations are sampled from the current population $\mathcal{P}_t$, with probability according to their fitness rank. They are first processed by the diagnostic LLM to produce a structural diagnosis (Section~\ref{sec:diagnosis}). Guided by this diagnosis, the generator LLM produces an offspring formulation that integrates complementary strengths or explores an orthogonal modeling direction. For mutation, a high-performing elite is selected. Similar to the crossover workflow, the diagnostic LLM analyzes the elite formulation and pinpoints specific refinement opportunities to explore its structural neighborhood. Guided by this targeted diagnosis, the generator LLM then implements directed refinements instead of blind perturbation to produce a mutated offspring.

All offspring candidates are subject to the evaluation and repair procedure described above. This process iterates for $T$ generations, to evolve the population toward increasingly solver-efficient formulations. The best formulation across all generations is returned as the final formulation $\hat{f}^\star$.

\subsection{Solver-Informed Diagnosis} \label{sec:diagnosis}

A key bottleneck of evolutionary search is that it overly relies on a scalar fitness signal to guide the evolution~\cite{romera2024mathematical,shi2026generalizable}. In our case, the runtime fitness can indicate \textit{whether} a formulation is slow, but provides little insight into \textit{why} it is slow, making the search space exploration highly sample-inefficient. 

To overcome this bottleneck, FormuEvo introduces a \textit{solver-informed diagnosis} mechanism. We note that modern MIP solvers are not mere black-box evaluators; instead, they expose rich, fine-grained internal statistics during optimization, such as presolving outcomes, relaxation quality, and branching dynamics (right of Fig.~\ref{fig:framework}). These statistics reflect important solver behaviors and reveal structural bottlenecks that a single runtime cannot capture. For example, a weak relaxation may indicate excessive fractional solutions that require strengthening, while a large branch-and-bound tree may suggest a highly symmetric structure.

To exploit this fine-grained feedback for targeted, solver-informed evolution, the diagnostic LLM takes the parent or elite formulations alongside their solver statistics to produce a structured verbal diagnosis before crossover or mutation. The diagnosis identifies major computational bottlenecks, attributes them to specific structural properties, and proposes targeted directions for improvement. For instance, if parent $f_i$ is compact but suffers from weak relaxations that induce excessive branching, while parent $f_j$ yields stronger bounds but incurs costly presolving, the diagnostic LLM may guide the generator LLM to selectively incorporate the bound-tightening constraints of $f_j$ where they matter most, while retaining the compact structure of $f_i$ elsewhere. This diagnosis functions as an interpretable verbal gradient, guiding the generator LLM toward principled, solver-informed structural improvement rather than undirected exploration.

\subsection{Structured Memory} \label{sec:memory}

We note that in long-horizon evolutionary search, LLMs are prone to redundant exploration, such as repeatedly revisiting failed modifications or forgetting effective sub-structures that can be directly retrieved for future refinements. To mitigate this issue, FormuEvo incorporates a structured memory library to abstract, organize, and store both successful and failed experiences into reusable modeling strategies for guiding future evolution steps.

\begin{figure}[!t]
\centering
\includegraphics[width=0.8\linewidth]{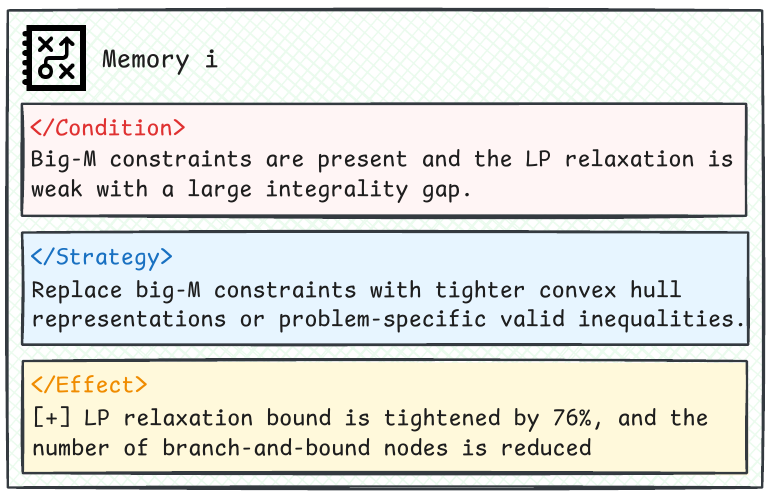}
\caption{Example of a structured memory entry.}
\label{fig:memory}
\end{figure}

Specifically, after each evaluation, the reflector LLM abstracts the modeling modifications and solver feedback between parent and offspring formulations into a structured memory entry with three fields: 1) a \textit{condition} that describes the problem context and formulation characteristics under which the strategy was applied; 2) a \textit{strategy} that captures the specific modeling decision or structural modification; and 3) an \textit{effect} that summarizes its observed impact on solver behavior and optimization performance. This structured format, as illustrated in Fig.~\ref{fig:memory}, enables precise retrieval and reuse of prior experience. During subsequent evolution, the diagnostic LLM retrieves relevant memory entries based on the current bottleneck conditions and incorporates them as additional context to enrich the diagnosis, leveraging prior experience to prune the search space and avoid known pitfalls.

While the memory efficiently accelerates the search within a single problem, we note that optimization problems often share common modeling logic and structural properties (e.g., routing logic, assignment constraints), even when their mathematical formulations may appear different. Therefore, upon completion of the evolutionary search, we introduce a distiller LLM to summarize the accumulated memory library into high-level problem-agnostic knowledge by removing instance-specific details. This distilled knowledge base not only enables FormuEvo to perform efficient zero-shot transfer when tasked with entirely unseen optimization problems, but can also serve as a lightweight yet high-quality reasoning resource for bootstrapping smaller LLMs with domain expertise distilled from a more powerful model (see Section~\ref{sec:transfer}).

\section{Experiments}

\subsection{Experiment Setup}

\paragraph{Benchmarks.} We evaluate FormuEvo on classical mixed-integer linear programming (MILP) and non-linear programming (MINLP) benchmarks, including: traveling salesman problem (TSP), job-shop scheduling problem (JSSP), bin packing problem (BPP), capacitated facility location problem (CFLP), and non-linear quadratic assignment problem (QAP). These problems span diverse optimization structures, covering routing, scheduling, packing, location, and assignment. To further assess the capability of FormuEvo beyond well-studied textbook settings, we additionally consider two challenging and less explored optimization tasks: neural network verification (NNV)~\cite{tjeng2018evaluating} and a mathematical challenge derived from IMO 2025 Problem 6~\cite{aops2025p6}, where human-designed formulation priors are limited or absent. Following prior practices~\cite{gleixner2021miplib}, instances for classical problems are categorized into three levels: Easy, Medium, and Hard. The Easy set consists of small instances used for training and validating in evolution, while the Medium and Hard sets contain larger instances held out to test the final performance. For the two novel optimization problems, we use Easy instances for evolution, and evaluate final performance on a separate set of Hard instances. More detailed problem descriptions and the corresponding instance splits are provided in Appendix~\ref{app:problem}.

\paragraph{Baselines.} We compare FormuEvo with two categories of baselines. First, we include both standard and state-of-the-art expert-designed formulations from the operations research literature, tracing the advancement of human expertise. For instance, TSP baselines range from the textbook MTZ formulation~\cite{miller1960integer}, to the stronger single-commodity flow (SCF) formulation~\cite{gavish1978travelling}, up to the MCF-RLT~\cite{balma2018strong}, which achieves the theoretically tightest known relaxation bound, representing the current limits of human design. We also compare with recent LLM-based modeling approaches, including ORLM~\cite{huang2025orlm} and StepORLM~\cite{zhou2026steporlm}, to demonstrate the gap between correctness-oriented generation and efficiency-oriented evolution. Finally, we include EvoCut~\cite{yazdani2025evocut}, an LLM-guided cutting-plane generation method, to highlight the difference between local formulation strengthening and our exploration of the broad symbolic space of MIP formulations. Additional introductions for the baseline formulations of each problem are also provided in Appendix~\ref{app:problem}.

\paragraph{Implementation.} 
All experiments are conducted in Python using Gurobi 10.0 as the downstream MIP solver, with a single thread and default parameters, on a server equipped with AMD EPYC 9654 processors. We use a population size of $N=8$ and evolve for $T=5$ generations, producing 8 offspring per generation via crossover and mutation with a mutation rate of 0.3. Memory-augmented generation is applied with probability 0.7 to avoid search overfitting toward previously explored regions. Each candidate is evaluated on 100 Easy instances, with fitness measured by SGM runtime with a 1-second shift~\cite{gleixner2021miplib}. Failed candidates receive one repair attempt. Operator prompts are provided in Appendix~\ref{app:implementation}.

FormuEvo uses GPT-5.4-mini as the backbone LLM (a comparison across different backbones is reported in Section~\ref{sec:backbone}). ORLM and StepORLM use their fine-tuned LLMs, and are allowed the same number of generation attempts as FormuEvo for a fair comparison. For EvoCut, we adopt the reported formulation from their original paper on problems covered by their benchmark, noting that EvoCut uses a stronger backbone model than ours. For problems outside their benchmark suite, we rerun EvoCut using the same backbone and experimental settings as FormuEvo.

\subsection{Main Results}
Tables~\ref{tab:classical}, \ref{tab:novel} present results on classical MILP and MINLP benchmarks and the two novel problems, respectively. We report three metrics: 1) ``Time'', SGM runtime over 100 test instances and 5 independent runs, with a per-instance time limit of 600 seconds; 2) ``Wins'', the number of instances on which a formulation achieves the best runtime among all formulations; and 3) ``Solved'', the number of instances solved to optimality within the time limit.

Overall, FormuEvo discovers formulations that outperform both expert-designed and LLM-based baselines, with acceleration of up to $5.5\times$ over the best baselines. The best discovered formulations are presented in Appendix~\ref{app:problem}. Moreover, FormuEvo attains the best runtime on the majority of instances across nearly all benchmarks, demonstrating robustness across diverse optimization structures. Beyond the numerical improvements, the results reveal several important insights.

\begin{table*}[!t]
\centering
\caption{Performance on classical MILP and MINLP problems.}\label{tab:classical}
\setlength{\tabcolsep}{4pt}
\resizebox{0.95\linewidth}{!}{
\begin{tabular}{ll|cc|cc|ccc}
\toprule
&& \multicolumn{2}{c|}{Easy} & \multicolumn{2}{c|}{Medium} & \multicolumn{3}{c}{Hard} \\
\midrule
& Formulation & Time$\downarrow$ & Wins$\uparrow$ & Time$\downarrow$ & Wins$\uparrow$ & Time$\downarrow$ & Wins$\uparrow$ & Solved$\uparrow$ \\
\midrule
\multirow{7}{*}{\rotatebox[origin=c]{90}{TSP}}
& MTZ{\footnotesize~\shortcite{miller1960integer}}      & 0.9578 & 0/100 & 6.1826 & 5/100 & 95.1315 & 0/100 & 62/100 \\
& SCF{\footnotesize~\shortcite{gavish1978travelling}}    & 0.3079 & 3/100 & 1.0065 & 10/100 & 8.3315 & 3/100 & \textbf{100/100} \\
& MCF-RLT{\footnotesize~\shortcite{balma2018strong}}   & 3.2729 & 0/100 & 84.1649 & 0/100 & 600.0570 & 0/100 & 0/100 \\
\cmidrule{2-9}
& ORLM{\footnotesize~\shortcite{huang2025orlm}}   & 0.9575 & 2/100 & 6.3270 & 0/100 & 92.5953 & 0/100 & 62/100 \\
& StepORLM{\footnotesize~\shortcite{zhou2026steporlm}}   & 1.7667 & 0/100 & 17.0081 & 0/100 & 173.8792 & 0/100 & 64/100 \\
& EvoCut{\footnotesize~\shortcite{yazdani2025evocut}}    & 0.5268 & 8/100 & 2.5013 & 6/100 & 13.8359 & 17/100 & 92/100 \\
\cmidrule{2-9}
& FormuEvo (Ours)    & \textbf{0.1410}{\scriptsize~(+54.2\%)} & \textbf{87/100} & \textbf{0.5640}{\scriptsize~(+44.0\%)} & \textbf{79/100} & \textbf{3.9469}{\scriptsize~(+52.6\%)} & \textbf{80/100} & \textbf{100/100} \\
\midrule
\midrule
\multirow{6}{*}{\rotatebox[origin=c]{90}{JSSP}}
& Disj.{\footnotesize~\shortcite{manne1960job}}         & 0.3264 & 19/100 & 1.8389 & 9/100 & 17.6653 & 25/100 & 98/100 \\
& Enh. Disj.{\footnotesize~\shortcite{ku2016mixed}}        & 0.4886 & 1/100 & 7.9405 & 0/100 & 158.1492 & 0/100 & 77/100 \\
\cmidrule{2-9}
& ORLM{\footnotesize~\shortcite{huang2025orlm}}  & 600.0012 & 0/100 & 600.0032 & 0/100 & 600.0019 & 0/100 & 0/100 \\
& StepORLM{\footnotesize~\shortcite{zhou2026steporlm}}  & 0.3040 & 28/100 & 1.7270 & 22/100 & 18.7024 & 20/100 & \textbf{99/100} \\
& EvoCut{\footnotesize~\shortcite{yazdani2025evocut}}   & 0.3005 & 19/100 & 1.7627 & 20/100 & 18.1329 & 14/100 & 98/100 \\
\cmidrule{2-9}
& FormuEvo (Ours)    & \textbf{0.2781}{\scriptsize~(+7.5\%)} & \textbf{33/100} & \textbf{1.5316}{\scriptsize~(+11.3\%)} & \textbf{49/100} & \textbf{15.4237}{\scriptsize~(+12.7\%)} & \textbf{41/100} & 98/100 \\
\midrule
\midrule
\multirow{7}{*}{\rotatebox[origin=c]{90}{BPP}}
& Kant.{\footnotesize~\shortcite{kantorovich1960mathematical}}       & 18.6428 & 0/100 & 191.9222 & 0/100 & 351.7605 & 0/100 & 28/100 \\
& AF{\footnotesize~\shortcite{valerio1999exact}}   & 0.2710 & 0/100 & 0.9930 & 1/100 & 4.9778 & 1/100 & \textbf{100/100} \\
& VPSolver{\footnotesize~\shortcite{brandao2016bin}}   & 0.0610 & \textbf{53/100} & 0.4064 & 40/100 & 1.4425 & 38/100 & \textbf{100/100} \\
\cmidrule{2-9}
& ORLM{\footnotesize~\shortcite{huang2025orlm}}   & 18.3614 & 0/100 & 287.7057 & 0/100 & 341.1659 & 0/100 & 24/100 \\
& StepORLM{\footnotesize~\shortcite{zhou2026steporlm}}   & 12.3430 & 0/100 & 117.0032 & 0/100 & 95.0336 & 1/100 & 42/100 \\
& EvoCut{\footnotesize~\shortcite{yazdani2025evocut}}     & 3.0279 & 0/100 & 26.7969 & 4/100 & 29.0496 & 5/100 & 60/100 \\
\cmidrule{2-9}
& FormuEvo (Ours)    & \textbf{0.0543}{\scriptsize~(+11.0\%)} & 47/100 & \textbf{0.3418}{\scriptsize~(+15.9\%)} & \textbf{55/100} & \textbf{0.8384}{\scriptsize~(+41.9\%)} & \textbf{55/100} & \textbf{100/100} \\
\midrule
\midrule
\multirow{7}{*}{\rotatebox[origin=c]{90}{CFLP}}
& Agg.{\footnotesize~\shortcite{balinski1965integer}}      & 2.7178 & 19/100 & 22.5520 & 8/100 & 87.3078 & 6/100 & 96/100 \\
& Disagg.{\footnotesize~\shortcite{leung1989valid}}   & 4.8283 & 0/100 & 31.4982 & 0/100 & 113.4126 & 1/100 & 95/100 \\
& Enh. Disagg.{\footnotesize~\shortcite{avella2021weak}}   & 5.0126 & 0/100 & 33.1480 & 0/100 & 113.5033 & 1/100 & 95/100 \\
\cmidrule{2-9}
& ORLM{\footnotesize~\shortcite{huang2025orlm}}   & 2.8439 & 0/100 & 24.0395 & 0/100 & 88.4319 & 2/100 & 95/100 \\
& StepORLM{\footnotesize~\shortcite{zhou2026steporlm}}   & 3.2663 & 1/100 & 26.8000 & 3/100 & 104.8201 & 3/100 & 94/100 \\
& EvoCut{\footnotesize~\shortcite{yazdani2025evocut}}    & 2.7698 & 5/100 & 17.1088 & 14/100 & 59.4309 & 16/100 & \textbf{100/100} \\
\cmidrule{2-9}
& FormuEvo (Ours)    & \textbf{1.6258}{\scriptsize~(+40.2\%)} & \textbf{75/100} & \textbf{11.6952}{\scriptsize~(+31.6\%)} & \textbf{75/100} & \textbf{44.3447}{\scriptsize~(+25.4\%)} & \textbf{71/100} & \textbf{100/100} \\
\midrule
\midrule
\multirow{9}{*}{\rotatebox[origin=c]{90}{QAP}}
& KB Quad.{\footnotesize~\shortcite{koopmans1957assignment}} & 9.9323 & 0/100 & 23.7309 & 0/100 & 538.7275 & 0/100 & 28/100 \\
& McC. Lin.{\footnotesize~\shortcite{mccormick1976computability}}  & 9.5458 & 0/100 & 67.9480 & 0/100 & 600.0062 & 0/100 & 0/100 \\
& AJ Lin.{\footnotesize~\shortcite{adams1994improved}}   & 1.2154 & 1/100 & 8.0087 & 0/100 & 237.7423 & 0/100 & 97/100 \\
& XY-KB Lin.{\footnotesize~\shortcite{xia2006new}} & 0.4044 & 28/100 & 1.6163 & 25/100 & 37.8213 & 28/100 & \textbf{100/100} \\ 
& IPQAPR{\footnotesize~\shortcite{zhang2010effective}}   & \textbf{0.3929} & 35/100 & 1.6883 & 26/100 & 40.2989 & 19/100 & \textbf{100/100} \\
\cmidrule{2-9}
& ORLM{\footnotesize~\shortcite{huang2025orlm}}\textsuperscript{*}   & - & - & - & - & - & - & - \\
& StepORLM{\footnotesize~\shortcite{zhou2026steporlm}}   & 9.7489 & 0/100 & 68.3794 & 0/100 & 600.0061 & 0/100 & 0/100 \\
& EvoCut{\footnotesize~\shortcite{yazdani2025evocut}}   & 10.1303 & 0/100 & 23.9718 & 0/100 & 538.5478 & 0/100 & 28/100 \\
\cmidrule{2-9}
& FormuEvo (Ours) & 0.3936{\scriptsize~(-0.2\%)} & \textbf{36/100} & \textbf{1.5648}{\scriptsize~(+3.2\%)} & \textbf{49/100} & \textbf{34.5034}{\scriptsize~(+8.8\%)} & \textbf{53/100} & \textbf{100/100} \\
\bottomrule
\end{tabular}
}
\begin{minipage}{0.95\linewidth}
\footnotesize \textsuperscript{*}ORLM failed to produce a correct formulation for QAP.
\end{minipage}
\end{table*}

First, FormuEvo successfully discovers formulations that outperform not only textbook models but also highly optimized expert-designed formulations, often by substantial margins. For instance, in TSP, while the human-designed MCF-RLT formulation has the theoretically tightest relaxation bound, it completely fails (0/100 solved) on Hard instances. This highlights a known but often overlooked fact: theoretical tightness does not necessarily translate to solver efficiency, as a heavily extended formulation can overwhelm the computation in MIP solvers. In contrast, FormuEvo is efficiency-oriented and guided by solver-informed statistics, enabling targeted optimization toward the practical computational objective of solvers.

Second, the results of ORLM and StepORLM further illustrate the gap between correctness and efficiency. Although they successfully generate correct and executable formulations on standard MILPs, their outputs largely remain naive textbook formulations, which are weak and scale poorly in practice. Moreover, both approaches fail to produce valid formulations for the novel NNV and IMO problems. This also underscores the limited generalization of fine-tuning approaches to more complex problems beyond their training distribution. See Appendix~\ref{app:discussion} for further discussion.

Third, EvoCut, which strengthens fixed formulations through LLM-generated cutting planes, yields moderate improvements over baseline formulations but remains fundamentally limited by the underlying model structure. For instance, on BPP, EvoCut improves upon standard formulations but remains significantly weaker than both VPSolver and FormuEvo, because the primary bottleneck of BPP lies in its assignment structure rather than LP relaxation strength. Similarly, on QAP, the standard quadratic formulation defines a Birkhoff polytope that already represents the convex hull~\cite{von1953certain}, and EvoCut fails to achieve any improvement through strengthening cuts. In contrast, FormuEvo breaks through this limitation by searching over the entire symbolic space of MIP formulations, enabling the discovery of structural reformulations such as alternative linearizations and variable extension that fundamentally alter the search landscape rather than merely modifying a fixed one.

\begin{table}[!t]
\centering
\caption{Performance on two novel MIP problems.}\label{tab:novel}
\setlength{\tabcolsep}{4pt}
\resizebox{0.95\linewidth}{!}{
\begin{tabular}{ll|ccc}
\toprule
& Formulation & Time$\downarrow$ & Wins$\uparrow$ & Solved$\uparrow$ \\
\midrule
\multirow{4}{*}{\rotatebox[origin=c]{90}{NNV}}
&  Std. Formu.{\footnotesize~\shortcite{tjeng2018evaluating}} & 69.5200 & 9/100 & 71/100 \\
&  ORLM / StepORLM  &  - & - &  - \\
&  EvoCut{\footnotesize~\shortcite{yazdani2025evocut}} & 67.6093 &  13/100  &  73/100 \\
\cmidrule{2-5}
& \raisebox{0.2\normalbaselineskip}{FormuEvo (Ours)} & \shortstack{\textbf{21.4139}\\{\scriptsize(+68.3\%)}\vspace{-0.3em}} & \raisebox{0.2\normalbaselineskip}{\textbf{78/100}} & \raisebox{0.2\normalbaselineskip}{\textbf{86/100}} \\
\midrule
\midrule
\multirow{4}{*}{\rotatebox[origin=c]{90}{IMO}}
&  Std. Formu.{\footnotesize~\shortcite{aops2025p6}} &  114.0912 &  0/4  &  3/4 \\
&  ORLM / StepORLM  &  - & - &  - \\
&  EvoCut{\footnotesize~\shortcite{yazdani2025evocut}} &  99.4400 &  0/4  &  3/4 \\
\cmidrule{2-5}
& \raisebox{0.2\normalbaselineskip}{FormuEvo (Ours)} & \shortstack{\textbf{17.9341}\\{\scriptsize(+82.0\%)}\vspace{-0.3em}} & \raisebox{0.2\normalbaselineskip}{\textbf{4/4}} & \raisebox{0.2\normalbaselineskip}{\textbf{4/4}} \\
\bottomrule
\end{tabular}
}
\end{table}

\begin{table}[!t]
\centering
\caption{Ablation study of FormuEvo components.}\label{tab:ablation}
\setlength{\tabcolsep}{6pt}
\resizebox{0.95\linewidth}{!}{
\begin{tabular}{ll|c|c|c}
\toprule
& Method &Easy & Medium & Hard \\
\midrule
\multirow{4}{*}{\rotatebox[origin=c]{90}{TSP}}
& Best Baseline & 0.3079 & 1.0065 & 8.3315 \\
& FormuEvo & \textbf{0.1410} & \textbf{0.5640} & \textbf{3.9469} \\
& w/o Memory & 0.1776 & 0.6102 & 4.6649 \\
& w/o Diagnosis & 0.2308 & 0.7350 & 6.5207 \\
\midrule
\midrule
\multirow{4}{*}{\rotatebox[origin=c]{90}{JSSP}}
& Best Baseline & 0.3005 & 1.7270 & 17.6653 \\
& FormuEvo & \textbf{0.2781} & \textbf{1.5316} & \textbf{15.4237} \\
& w/o Memory & 0.2846 & 1.5698 & 16.3223 \\
& w/o Diagnosis & 0.3048 & 1.7648 & 18.0159 \\
\bottomrule
\end{tabular}
}
\end{table}

\subsection{Transfer Performance} \label{sec:transfer}

\begin{figure}[!t]
\centering
\includegraphics[width=\linewidth]{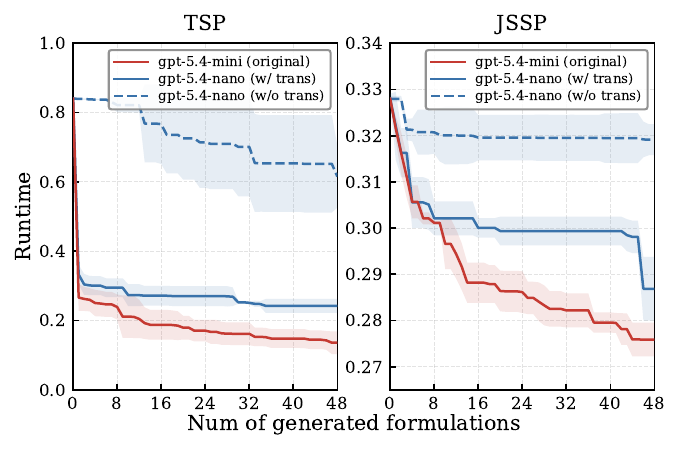}
\vspace{-10pt}
\caption{Transfer performance on smaller LLMs.}
\label{fig:transfer}
\end{figure}

We evaluate the transferability of distilled knowledge to unseen problems and smaller LLMs. Fig.~\ref{fig:transfer} shows evolution trajectories using a smaller LLM (GPT-5.4-nano), with distilled knowledge aggregated from all problems except the target. Without transfer, GPT-5.4-nano exhibits substantially slower convergence and higher final runtime, struggling to move beyond the baseline formulations. In contrast, when augmented with the distilled knowledge, GPT-5.4-nano significantly narrows the gap, achieving convergence trajectories and final performance close to the larger GPT-5.4-mini model. These results suggest that the distilled knowledge can serve as a lightweight but effective reasoning source, compensating for the capacity of smaller models by providing structured domain expertise that would otherwise be difficult to infer independently, which offers a practical pathway toward strong MIP formulation discovery at lower costs.

\subsection{Ablation Study on Key Components}

We analyze the contributions of two key components: solver-informed diagnosis and memory, by comparing FormuEvo against two ablated variants. As shown in Table~\ref{tab:ablation}, removing either component leads to clear performance degradation, indicating that the diagnosis mechanism plays a crucial role in providing the directional signal for FormuEvo, while the memory reuse also facilitates the evolution by improving the search efficiency.

\subsection{Impact of Different LLM Backbones} \label{sec:backbone}

\begin{table}[!t]
\centering
\caption{Impact of different backbone LLMs.}\label{tab:backbone}
\setlength{\tabcolsep}{4pt}
\resizebox{\linewidth}{!}{
\begin{tabular}{ll|c|c|c|c}
\toprule
& Method & Cost & Easy & Medium & Hard \\
\midrule
\multirow{4}{*}{\rotatebox[origin=c]{90}{TSP}}
& Best Baseline & - & 0.3079 & 1.0065 & 8.3315 \\
& GPT-5.4-mini & $\sim \mathdollar$2 & 0.1410 & 0.5640 & \textbf{3.9469} \\
& Claude-Sonnet-4.6 & $\sim \mathdollar$10 & \textbf{0.1017} & \textbf{0.4777} & 4.0148 \\
& DeepSeek-V4-Flash & $\sim \mathdollar$0.5 & 0.1498 & 0.6283 & 5.3068 \\
\midrule
\midrule
\multirow{4}{*}{\rotatebox[origin=c]{90}{JSSP}}
& Best Baseline & - & 0.3005 & 1.7270 & 17.6653 \\
& GPT-5.4-mini & $\sim \mathdollar$3 & \textbf{0.2781} & \textbf{1.5316} & \textbf{15.4237} \\
& Claude-Sonnet-4.6 & $\sim \mathdollar$16 & 0.2808 & 1.5319 & 15.8340 \\
& DeepSeek-V4-Flash & $\sim \mathdollar$0.8 & 0.3050 & 1.6399 & 16.1327 \\
\bottomrule
\end{tabular}
}
\end{table}

We further evaluate FormuEvo with different backbone LLMs (Table~\ref{tab:backbone}). All three consistently outperform the best baselines on both problems, with only modest differences in formulation quality and generalization, where no backbone model consistently dominates across all settings. These results indicate that the gains stem primarily from the evolutionary framework and are robust to the choice of backbone. We adopt GPT-5.4-mini as the default because it offers the best overall trade-off between performance and API cost.

Additional analyses on different downstream solvers, backbone LLMs, and evolutionary hyperparameters are provided in Appendix~\ref{app:robustness}~-~\ref{app:hyperparameters}.

\section{Conclusions}

We presented FormuEvo, an LLM-guided evolutionary framework for automatically discovering solver-efficient MIP formulations, which incorporates solver-informed diagnosis and structured memory for directed and systematic improvement. Experiments show that FormuEvo successfully discovers formulations that significantly outperform state-of-the-art baselines, accelerating modern MIP solvers by up to $5.5\times$. Notably, FormuEvo is transferable to unseen problems and smaller LLMs by distilling a generalizable knowledge base, demonstrating its generalizability and applicability.

\section*{Limitations}

FormuEvo focuses on discovering static MIP formulations that can be directly solved by off-the-shelf general-purpose MIP solvers, which represents the most commonly adopted approach in real-world practice. However, for many large-scale or highly complex optimization problems, advanced problem-specific solution approaches in operations research often rely on dynamic relaxation or decomposition algorithms, such as column generation, which iteratively adds variables by solving a pricing sub-problem, and Benders decomposition, which relaxes the original problem and then alternates between a master problem and sub-problems to progressively tighten the formulation. These methods typically require iterative interactions between master problems and sub-problems, along with dynamic generation of variables and constraints during the solving process. In such approaches, \textit{formulation} design and \textit{algorithm} development are inherently coupled: the effectiveness of the formulation depends on the decomposition strategy, while the decomposition procedure itself is tightly determined by the underlying formulation structure. Extending FormuEvo beyond static formulations to jointly evolve reformulations and decomposition-based solution algorithms is a promising but substantially more challenging direction. We leave this extension for future work.

\section*{Acknowledgments}

This work was done during the first author's internship at Singapore Management University. This work is supported in part by the BNRist Project (No:~BNR2024TD03003), the National Natural Science Foundation of China (No:~42327901), and the Program of China Scholarship Council (No:~202506210230). This research is supported by the National Research Foundation Singapore under the AI Singapore Programme (AISG Award No:~AISG3-RPGV-2025-017). We would also like to thank the AI4Opt group for helpful discussions, and the reviewers for their valuable feedback.

\bibliography{reference}

\clearpage
\newpage
\newpage

\appendix

\section{Problems and Formulations}\label{app:problem}

This appendix details the benchmark problems used in our experiments, including the descriptions, standard formulations, baselines, and the best formulations discovered by FormuEvo. Detailed instance splits in the experiments are shown in Table~\ref{tab:split}.

\begin{table}[h]
\centering
\caption{Instance splits for experiments.}\label{tab:split}
\setlength{\tabcolsep}{4pt}
\resizebox{0.9\linewidth}{!}{
\begin{tabular}{c|c|c}
\toprule
Problem & Category & Configuration \\
\midrule
\multirow{3}{*}{TSP}
& Easy & 30 cities\\
& Medium & 50 cities\\
& Hard & 100 cities\\
\midrule
\multirow{3}{*}{JSSP}
& Easy & 8 jobs, 8 machines\\
& Medium & 10 jobs, 10 machines\\
& Hard & 12 jobs, 12 machines\\
\midrule
\multirow{3}{*}{BPP}
& Easy & 100 items, 150 capacity\\
& Medium & 150 items, 300 capacity\\
& Hard & 200 items, 400 capacity\\
\midrule
\multirow{3}{*}{CFLP}
& Easy & 100 facilities, 200 customers\\
& Medium & 200 facilities, 400 customers\\
& Hard & 300 facilities, 600 customers\\
\midrule
\multirow{3}{*}{QAP}
& Easy & 8 facilities / locations\\
& Medium & 9 facilities / locations\\
& Hard & 10 facilities / locations\\
\midrule
\multirow{2}{*}{NNV}
& Easy & MLP dim [20, 80, 10]\\
& Hard & MLP dim [40, 160, 20]\\
\midrule
\multirow{2}{*}{IMO}
& Easy & Grid size $\{3, 4, 5, 6\}$\\
& Hard & Grid size $\{7, 8, 9, 10\}$\\
\bottomrule
\end{tabular}
}
\vspace{-5pt}
\end{table}

\subsection{Traveling Salesman Problem (TSP)}

\subsubsection*{Problem description}

The traveling salesman problem aims to determine an optimal Hamiltonian cycle (route) to visit a set of $n$ cities exactly once and return to the starting city. The problem is defined on a complete graph where each pair of cities $(i, j)$ is associated with a travel distance (or cost) $c_{ij}$. The decision involves sequencing the visits to ensure that every city is entered exactly once and exited exactly once. The selected route must form a single continuous tour covering all $n$ cities. Any disconnected, smaller loops (subtours) that do not include all cities are strictly prohibited. The objective is to minimize the total travel distance of the complete tour.

\subsubsection*{Standard MIP formulation}
\begin{itemize}[leftmargin=*, itemsep=0pt, topsep=0pt, parsep=0pt]
    \item $V=\{1,2,\dots,n\}$: the set of $n$ cities.
    \item $C=V\setminus\{1\}$: the set of cities except city $1$.
    \item $c_{ij}$: travel distance from city $i$ to city $j$.
    \item $x_{ij}$: decision variable that equals 1 if and only if the route travels directly from node $i$ to node $j$.
    \item $u_{i}$: auxiliary ordering variable used for subtour elimination, where $u_1=1$.
\end{itemize}
Below is the standard MTZ formulation for TSP:
\begin{align*}
\min \; & \sum_{i=1}^{n} \sum_{j=1,\ j \neq i}^{n} c_{ij} x_{ij} \\
\mathrm{s.t.} \; & \hspace{-0.5em} \sum_{j=1,\ j \neq i}^{n} x_{ij} = 1, \quad \forall i \in V, \\
& \hspace{-0.5em} \sum_{i=1,\ i \neq j}^{n} x_{ij} = 1, \quad \forall j \in V, \\
& \hspace{-0.5em} u_i - u_j \le (n - 1)(1 - x_{ij}),\ \forall i, j \in C, \ i \neq j, \\
& \hspace{-0.5em} x_{ij} \in \{0, 1\}, \quad \forall i, j \in V,\ i \neq j, \\
& \hspace{-0.5em} 2 \le u_i \le n, \quad \forall i \in C.
\end{align*}
\subsubsection*{Baseline formulations}
\begin{itemize}[leftmargin=*, itemsep=2pt, topsep=0pt, parsep=0pt]
    \item MTZ: The standard formulation using continuous auxiliary sequencing variables to enforce a strictly increasing visit order for subtour elimination, as described above.
    \item SCF: A flow-based formulation that eliminates subtours by tracking a single continuous flow where the depot dispatches $n-1$ units and each subsequent city consumes exactly one unit.
    \item MCF-RLT: An advanced flow-based formulation using reformulation-linearization technique, which linearizes higher-order products of flow and precedence variables, achieving the theoretically tightest known relaxation bound.
    \item ORLM: Generated by ORLM, same as MTZ.
    \item StepORLM: Generated by StepORLM, based on MTZ with explicit self-loop elimination.
    \item EvoCut: A strengthened MTZ with lifted ordering constraints, envelope bounds, and elimination of 2-cycles and depot-associated short cycles.
\end{itemize}

\subsubsection*{FormuEvo formulation}
\begin{itemize}[leftmargin=*, itemsep=0pt, topsep=0pt, parsep=0pt]
    \item $\bar{c}_{ij}=c_{ij}+c_{ji}$: bidirectional distance of $(i, j)$.
    \item $f_{ij}$: non-negative flow variable on arc $(i,j)$, defined for $i,j\in C$, $i\neq j$.
\end{itemize}
The FormuEvo formulation preserves the global connectivity guarantee of SCF formulation while selectively incorporating small lifted DFJ subtour elimination cuts as local strengthening constraints:
\begin{align*}
\min\; & \sum_{i=1}^{n} \sum_{j=1,\ j \neq i}^{n} c_{ij} x_{ij} \\
\mathrm{s.t.}\; & \sum_{j=1,\ j \neq i}^{n} x_{ij} = 1, \quad \forall i \in V, \\
& \sum_{i=1,\ i \neq j}^{n} x_{ij} = 1, \quad \forall j \in V, \\
& x_{ij} + x_{ji} \le 1, \quad \forall i, j \in V,\ i < j, \\
& \sum_{i\in S}\sum_{j\in S,\ j\neq i}x_{ij}\le |S|-1, \quad \forall S \in \mathcal{F}, \\
& (n-1) \cdot x_{1j} + \hspace{-0.05cm} \sum_{i \in C \setminus \{j\}} f_{ij}
   - \hspace{-0.05cm} \sum_{k \in C \setminus \{j\}} f_{jk} = 1,\\
& \hspace{5.7cm} \ \forall j \in C, \\
& x_{ij} \le f_{ij} \le (n-2)\cdot x_{ij}, \quad \forall i, j \in C,\ i \neq j, \\
& x_{ij} \in \{0, 1\}, \quad \forall i, j \in V,\ i \neq j, \\
& f_{ij} \ge 0, \quad \forall i, j \in C,\ i \neq j.
\end{align*}
The subtour elimination family $\mathcal{F}$ is defined as the set of all unique collection consisting of:
\begin{itemize}[leftmargin=*, itemsep=2pt, topsep=2pt, parsep=0pt]
    \item Depot $3$-cuts: $\{1,a,b\}$, $\forall a,b \in C,\ a < b$.
    \item Local $3$-cuts: For each $i \in C$, let $N_K(i)$ be the $K = \min(4, n-2)$ nearest neighbors of $i$ w.r.t.\ $\bar{c}_{ij}$. For every pair $\{a,b\} \subseteq N_K(i)$, add the set $\{i,a,b\}$ if $|\{i,a,b\}|<n$.
    \item Depot $4$-cuts (if $n>4$): Let $N_K(1)$ be the $K = \min(6, n-1)$ nearest neighbors of the depot w.r.t.\ $\bar{c}_{1j}$. For every triple $\{a,b,c\} \subseteq N_K(1)$, add the set $\{1,a,b,c\}$ if $|\{1,a,b,c\}|<n$.
    \item Local $4$-cuts (if $n>4$): For each $i \in C$ and every triple $\{a,b,c\} \subseteq N_K(i),\, K = \min(4, n-2)$, add the set $\{i,a,b,c\}$ if $|\{i,a,b,c\}|<n$.
\end{itemize}

\subsection{Job-Shop Scheduling Problem (JSSP)}

\subsubsection*{Problem description}

The job-shop scheduling problem aims to determine the optimal scheduling of a set of $n$ jobs on a set of $m$ machines. Each job $i$ consists of $m$ operations, where each operation must be processed on a distinct machine. The processing order of these operations is predefined for each job, representing its technological order. Each machine can process at most one operation at a time, and preemption is strictly prohibited. The decision involves determining the exact start time of each operation such that both machine capacity constraints and job precedence constraints are satisfied. The objective is to minimize the makespan, defined as the completion time of the last operation across all jobs.

\subsubsection*{Standard MIP formulation}
\begin{itemize}[leftmargin=*, itemsep=0pt, topsep=0pt, parsep=0pt]
    \item $J=\{1,\dots,n\}$: the set of $n$ jobs.
    \item $M=\{1,\dots,m\}$: the set of $m$ machines (also the number of operations per job).
    \item $\mu_{ij} \in M$: the machine that processes the $j$-th operation of job $i$.
    \item $\pi_i(k) \in M$: the index of the operation in job $i$ that executes on machine $k$, i.e., $\mu_{i,\pi_i(k)}=k$.
    \item $p_{ij}$: processing time of $j$-th operation of job $i$.
    \item $z_{i_1 i_2 k}$: binary variable that equals 1 if and only if job $i_1$ is processed before job $i_2$ on machine $k$.
    \item $s_{ij}$: decision variable representing the start time of the $j$-th operation of job $i$.
    \item $C_{\max}$: decision variable representing makespan.
    \item $L$: a large enough constant, set to $\sum_{i\in J,\, j\in M}p_{ij}$.
\end{itemize}
Below is the disjunctive formulation for JSSP:
\begin{align*}
\min \; & C_{\max} \\
\mathrm{s.t.}\; & s_{i,j+1} \ge s_{ij} + p_{ij},\ \forall i \in J,\ \forall j \in M \setminus \{m\},  \\
& s_{i_1,\pi_{i_1}(k)} + p_{i_1,\pi_{i_1}(k)} \le s_{i_2,\pi_{i_2}(k)} + \\
& L\cdot(1 - z_{i_1 i_2 k}),\, \forall i_1,i_2 \in J,\ i_1 < i_2,\ k \in M, \\
& s_{i_2,\pi_{i_2}(k)} + p_{i_2,\pi_{i_2}(k)} \le s_{i_1,\pi_{i_1}(k)} + L\cdot z_{i_1 i_2 k}, \\
& \hspace{2.5cm}\forall i_1,i_2 \in J,\ i_1 < i_2,\ k \in M, \\
& C_{\max} \ge s_{ij} + p_{ij},\quad \forall i \in J,\ j \in M, \\
& z_{i_1 i_2 k} \in \{0, 1\},\, \forall i_1,i_2 \in J,\ i_1 < i_2,\ k \in M. \\
& s_{ij} \ge 0,\quad \forall i\in J,\ j\in M, \\
& C_{\max}\ge0.
\end{align*}

\subsubsection*{Baseline formulations}
\begin{itemize}[leftmargin=*, itemsep=2pt, topsep=0pt, parsep=0pt]
    \item Disj.: The standard disjunctive formulation using big-$M$ constraints for non-overlapping requirements, as described above.
    \item Enh. Disj.: An enhanced disjunctive formulation that replaces standard big-$M$ inequalities with continuous surplus variables to explicitly track and bound the idle time between operations.
    \item ORLM: Generated by ORLM, a time-indexed formulation that enforces the precedence and non-overlapping requirements using time constraints.
    \item StepORLM: Generated by StepORLM, same as the disjunctive formulation.
    \item EvoCut: A strengthened disjunctive formulation that tightens the linear relaxation with static valid inequalities derived from maximum job lengths and machine-level critical-path bounds.
\end{itemize}
\subsubsection*{FormuEvo formulation}
\begin{itemize}[leftmargin=*, itemsep=0pt, topsep=0pt, parsep=0pt]
    \item $\underline{s}_{ij}=\sum_{l=1}^{j-1} p_{il}$: lower bound on $s_{ij}$
    \item $\sigma_{ij}=\sum_{l=j}^{m} p_{il}$: tail processing time from the $j$-th operation of job $i$ onward.
    \item $\bar{s}_{ij}=\max(\underline{s}_{ij},\,\hat{C}-\sigma_{ij})$: upper bound on $s_{ij}$, where $\hat{C}$ is a heuristic makespan.
    \item $\mu_i = (\mu_{i1},\dots,\mu_{im})$: the machine assignment vector of job $i$.
    \item $p_i = (p_{i1},\dots,p_{im})$: the processing time vector of job $i$.
    \item $\omega_j=m-j+1$: symmetry-breaking weight for operation $j$.
    \item $y_{i_1 i_2 k}$: binary disjunctive variable on machine $k$, defined only for \emph{free} pairs (pairs whose ordering cannot be inferred from bounds).
    \item $L_{a b k}=\bar{s}_{a,\pi_{a}(k)}+p_{a,\pi_{a}(k)}-\underline{s}_{b,\pi_{b}(k)}$: pair-specific tight big-$M$ for precedence $a \to b$.
\end{itemize}
The FormuEvo formulation strengthens the disjunctive model through tight variable bounds, tail-based makespan constraints, symmetry-breaking constraints, and bound-driven fixing of resolvable disjunctions to reduce binary variables.
\begin{align*}
\min \; & C_{\max} \\
\mathrm{s.t.}\; & s_{i,j+1} \ge s_{i,j} + p_{ij}, \; \forall i \in J,\ \forall j \in M\setminus\{m\}, \\
& C_{\max} \ge s_{i,j} + \sigma_{ij}, \quad \forall i \in J,\ \forall j \in M, \\
& \sum_{j=1}^{m} \omega_j\, s_{i_1,j} \le \sum_{j=1}^{m} \omega_j\, s_{i_2,j}, \\
& \hspace{1.0cm} \forall\, i_1,i_2 \in J \text{ with }\mu_{i_1}=\mu_{i_2},\ p_{i_1}=p_{i_2}, \\
& s_{i_1,\pi_{i_1}(k)} + p_{i_1,\pi_{i_1}(k)} \le s_{i_2,\pi_{i_2}(k)}, \\
& \hspace{3.1cm} \forall k \in M,\ \forall (i_1,i_2) \in \mathcal{P}_k, \\
& s_{i_1,\pi_{i_1}(k)} + p_{i_1,\pi_{i_1}(k)} \le s_{i_2,\pi_{i_2}(k)} + \\
& L_{i_1 i_2 k}\cdot(1-y_{i_1 i_2 k}), \forall k \in M,\ \forall (i_1,i_2) \in \mathcal{Y}_k, \\
& s_{i_2,\pi_{i_2}(k)} + p_{i_2,\pi_{i_2}(k)} \le s_{i_1,\pi_{i_1}(k)} + \\
& \hspace{0.9cm} L_{i_2 i_1 k}\cdot y_{i_1 i_2 k}, \forall k \in M,\ \forall (i_1,i_2) \in \mathcal{Y}_k, \\
& \underline{s}_{ij} \le s_{ij} \le \bar{s}_{ij}, \quad \forall i \in J,\ \forall j \in M, \\
& y_{i_1 i_2 k} \in \{0,1\}, \quad \forall k \in M,\ \forall (i_1,i_2) \in \mathcal{Y}_k.
\end{align*}
where $\mathcal{P}_k$ and $\mathcal{Y}_k$ are defined as follows. For each machine $k$, consider the operations processed on $k$ and sort them lexicographically by $(\underline{s}_{ij}+p_{ij},\,\bar{s}_{ij})$. For every ordered pair $(i_1,j_1)$ and $(i_2,j_2)$ in this order, if $\underline{s}_{i_1,j_1}+p_{i_1,j_1}>\bar{s}_{i_2,j_2}$,
then $(i_2,i_1)$ is added to a fixed precedence relation. Its transitive closure defines a partial order on machine $k$. Pairs resolved by this order are placed in $\mathcal{P}_k$, while the remaining unresolved pairs form $\mathcal{Y}_k$ and retain the binary big-$M$ disjunction. The heuristic makespan $\hat{C}$ is initialized from the best schedule produced by constructive heuristics, including serial list scheduling and priority dispatch rules.

\subsection{Bin Packing Problem (BPP)}

\subsubsection*{Problem description}
The bin packing problem aims to determine the optimal assignment of a set of $n$ discrete items into a set of identical bins. Each bin is homogeneous with a fixed, deterministic maximum weight capacity $W$. Each item $i$ has a predefined weight $w_i$, and all item weights are assumed to be positive integers. The packing decision involves determining exactly which bin each item is placed into. Every item must be assigned to exactly one bin. For any bin that is used, the cumulative weight of all items assigned to it cannot exceed $W$. Sufficient bins are assumed to be available, bounded by a maximum possible number $m$ (typically $m=n$). The objective is to minimize the total number of bins actively used to accommodate all $n$ items.

\subsubsection*{Standard MIP formulation}
\paragraph{Notation}
\begin{itemize}[leftmargin=*, itemsep=0pt, topsep=0pt, parsep=0pt]
    \item $I=\{1,2,\dots,n\}$: the set of $n$ items.
    \item $B=\{1,2,\dots,m\}$: the set of $m$ available bins.
    \item $w_i$: weight of item $i$.
    \item $W$: maximum weight capacity of each bin.
    \item $x_{ij}$: binary variable that equals $1$ if and only if item $i$ is placed into bin $j$.
    \item $y_j$: binary variable that equals $1$ if and only if bin $j$ is actively used.
\end{itemize}
Below is the Kantorovich formulation for BPP:
\begin{align*}
\min \; & \sum_{j=1}^{m} y_j \\
\mathrm{s.t.} \; & \sum_{j=1}^{m} x_{ij} = 1, \quad \forall i \in I, \\
& \sum_{i=1}^{n} w_i x_{ij} \le W y_j, \quad \forall j \in B, \\
& x_{ij} \in \{0, 1\}, \quad \forall i \in I,\ \forall j \in B, \\
& y_j \in \{0, 1\}, \quad \forall j \in B.
\end{align*}
\subsubsection*{Baseline formulations}
\begin{itemize}[leftmargin=*, itemsep=2pt, topsep=0pt, parsep=0pt]
    \item Kant.: The Kantorovich assignment formulation, as described above.
    \item AF: A pseudo-polynomial formulation that reformulates BPP as a flow network problem.
    \item VPSolver: An AF-style formulation with graph compression and side constraints.
    \item ORLM: Generated by ORLM, same as Kant.
    \item StepORLM: Generated by StepORLM, based on Kantorovich formulation with symmetry-breaking constraints and objective bounds.
    \item EvoCut: A strengthened Kantorovich variant that breaks symmetry via lexicographical ordering inequalities, continuous weight bounds, and index-based assignment restrictions for sorted items.
\end{itemize}

\subsubsection*{FormuEvo formulation}
\paragraph{Notation}
\begin{itemize}[leftmargin=*, itemsep=0pt, topsep=0pt, parsep=0pt]
    \item $g=\gcd(W,w_1,\dots,w_n)$: greatest common divisor (GCD) of $W$ and all weights $w_i$.
    \item $\mathcal{W} = \{\bar{w}_1,\dots,\bar{w}_K\}$: set of distinct item weights after GCD scaling ($W$ and $w_i$ are divided by $g$).
    \item $b_w$: the demand quantity of items with scaled weight $w \in \mathcal{W}$.
    \item $B_0$: the number of items whose weight equals the capacity $W$, which exactly fill a bin, and are pre-packed and excluded from the problem.
    \item $\mathcal{S} = \{0 = s_0 < s_1 < \dots < s_T \le W\}$: the sorted set of valid cumulative bin loads (nodes in the flow network).
    \item $x_{w,s}$: item flow variable representing the number of items of weight $w$ placed into bins that currently have an accumulated load of state $s$.
    \item $\ell_t$: loss flow variable representing the unused bin capacity between states $s_t$ and $s_{t+1}$.
    \item $z$: variable representing the number of bins required to pack all items lighter than $W$.
\end{itemize}
The FormuEvo formulation constructs a directed acyclic graph where valid residual capacities are compressed via subset-sum filtering. Loss arcs ($\ell_t$) ensure flow conservation without requiring explicit bin indexing, while $z$ and $\ell_t$ are naturally integers.
\begin{align*}
\min \; &\; B_0 + z \\
\mathrm{s.t.} \; & \sum_{w:\, w \in \mathcal{S}} x_{w,0} = z, \\
& \hspace{-0.2cm} \sum_{w:\, s_t-w \in \mathcal{S}} \hspace{-0.1cm} x_{w,\,s_t-w} + \ell_{t-1}
   = \hspace{-0.3cm} \sum_{w:\, s_t+w \in \mathcal{S}}\hspace{-0.1cm} x_{w,s_t} + \ell_t, \\
& \hspace{4.1cm} \forall t = 1,\dots,T - 1, \\
& \hspace{-0.2cm}\sum_{w:\, W-w \in \mathcal{S}} \hspace{-0.1cm} x_{w,\,W-w} + \ell_{T-1} = z, \\
& \hspace{-0.2cm} \sum_{s:\, s+w \in \mathcal{S}} \hspace{-0.1cm} x_{w,s} = b_w, \quad \forall w \in \mathcal{W}, \\
& \underline{z} \le z \le \overline{z}, \\
& x_{w,s} \in \mathbb{Z}_{\ge 0}, \quad \forall w \in \mathcal{W},\ s \in \mathcal{S},\ s+w \in \mathcal{S}, \\
& \ell_t \ge 0, \quad \forall t = 1,\dots,T - 1.
\end{align*}
where $\underline{z},\overline{z}$ are valid lower and upper bounds obtained from capacity-based bounds and a first-fit decreasing heuristic. The state set $\mathcal{S}$ is generated by subset-sum reachability over the scaled item sizes. Starting from 0, every reachable load $s+w$ with $s\in\mathcal{S}$ and $w\in\mathcal{W}$ is iteratively added whenever $s+w\le W$. Only transitions between reachable states are retained in the compressed network.

\subsection{Capacitated Facility Location Problem (CFLP)}

\subsubsection*{Problem description}

The capacitated facility location problem aims to determine the optimal subset of facilities to open from a set of $m$ candidate locations and the corresponding shipment quantities to satisfy the demands of $n$ customers. Each facility $i$ incurs a fixed opening cost $f_i$ and has a maximum supply capacity $C_i$. Each customer $j$ has a demand $d_j$ that must be fully met. It is assumed that the total available capacity is sufficient to satisfy total demand. Shipping one unit from location $i$ to customer $j$ incurs a transportation cost $c_{ij}$, and a facility can supply customers only if it is opened. The decision involves selecting which facilities to open and determining the exact shipment quantity from each opened facility to each customer such that all demands are met without exceeding any facility's capacity. The objective is to minimize the total cost, comprising the sum of fixed opening costs and variable transportation costs.

\subsubsection*{Standard MIP formulation}
\paragraph{Notation}
\begin{itemize}[leftmargin=*, itemsep=0pt, topsep=0pt, parsep=0pt]
    \item $I=\{1,2,\dots,m\}$: the set of $m$ locations.
    \item $J=\{1,2,\dots,n\}$: the set of $n$ customers.
    \item $f_i$: fixed cost of opening facility $i$.
    \item $C_i$: maximum supply capacity of facility $i$.
    \item $d_j$: demand of customer $j$.
    \item $c_{ij}$: unit transportation cost from facility $i$ to customer $j$.
    \item $y_i$: binary variable that equals $1$ if and only if facility $i$ is opened.
    \item $x_{ij}$: continuous variable representing the shipment quantity from facility $i$ to customer $j$.
\end{itemize}
The standard MIP formulation is as follows.
\begin{align*}
\min \; & \sum_{i=1}^{m} f_i y_i + \sum_{i=1}^{m} \sum_{j=1}^{n} c_{ij} x_{ij} \\
\mathrm{s.t.} \; & \sum_{i=1}^{m} x_{ij} = d_j, \quad \forall j \in J, \\
& \sum_{j=1}^{n} x_{ij} \le C_i y_i, \quad \forall i \in I, \\
& y_i \in \{0, 1\}, \quad \forall i \in I, \\
& x_{ij} \ge 0, \quad \forall i \in I,\ \forall j \in J.
\end{align*}

\subsubsection*{Baseline formulations}
\begin{itemize}[leftmargin=*, itemsep=2pt, topsep=0pt, parsep=0pt]
    \item Agg.: The aggregated formulation using facility-level capacity constraints, as described above.
    \item Disagg.: A disaggregated formulation that replaces the capacity constraint with per-customer linking inequalities for a tighter LP relaxation.
    \item ORLM: Generated by ORLM, same as the aggregated formulation.
    \item StepORLM: Generated by StepORLM, based on the aggregated formulation with explicitly decomposed demand constraints.
    \item EvoCut: A strengthened disaggregated formulation with tightened variable upper bounds, single-customer coverage inequalities, global capacity bounds, minimum-open-facility bounds, and subset coverage cuts for the largest customers.
\end{itemize}

\subsubsection*{FormuEvo formulation}
\paragraph{Notation}
\begin{itemize}[leftmargin=*, itemsep=0pt, topsep=0pt, parsep=0pt]
    \item $u_{ij} = \min(C_i,\, d_j)$: an upper bound on $x_{ij}$.
    \item $\mathcal{L}_j = \{i \in I \mid C_i \ge d_j\}$: set of large facilities whose individual capacity can satisfy demand $d_j$.
\end{itemize}
FormuEvo tightens the formulation through per-arc upper bounds, global and per-customer cover inequalities, minimum-open-count constraints, and symmetry breaking based on dominance.
\begin{align*}
\min \; & \sum_{i=1}^{m} f_i y_i + \sum_{i=1}^{m} \sum_{j=1}^{n} c_{ij} x_{ij} \\
\mathrm{s.t.} \; & \sum_{i=1}^{m} x_{ij} = d_j, \quad \forall j \in J, \\
& \sum_{j=1}^{n} x_{ij} \le C_i y_i, \quad \forall i \in I, \\
& \sum_{i=1}^{m} C_i y_i \ge \sum_{j=1}^{n} d_j, \\
& \sum_{i \in \mathcal{L}_j} C_i y_i \ge d_j - \hspace{-0.3cm} \sum_{i \in I \setminus \mathcal{L}_j} \hspace{-0.1cm} \min(C_i, d_j),\, \forall j \in J, \\
& \sum_{i=1}^{m} y_i \ge \Bigl\lceil \frac{\sum_{j=1}^{n} d_j}{\max_{i\in I} C_i} \Bigr\rceil, \\
& y_k \le y_i, \quad \forall\, i,k \in I \text{ with } C_i \ge C_k,\ f_i \le f_k,\ \\
& \hspace{3.6cm} \text{and } c_{ij} \le c_{kj}, \forall j \in J, \\
& 0 \le x_{ij} \le u_{ij}, \quad \forall i \in I,\ \forall j \in J, \\
& y_i \in \{0, 1\}, \quad \forall i \in I.
\end{align*}

\subsection{Quadratic Assignment Problem (QAP)}

\subsubsection*{Problem description}

The quadratic assignment problem aims to determine the optimal one-to-one assignment of a set of $n$ facilities to a set of $n$ distinct locations. Each facility must be assigned to exactly one location, and each location must be occupied by exactly one facility, establishing a strict bijective mapping. There is a deterministic flow $f_{ij}$ between every pair of facilities $i$ and $j$, and a deterministic distance $d_{kl}$ between every pair of locations $k$ and $l$. The decision involves determining a bijective assignment between facilities and locations. The objective is to minimize the total interaction cost, computed as the sum of flow–distance products induced by all assigned facility pairs.

\subsubsection*{Standard MIP formulation}
\paragraph{Notation}
\begin{itemize}[leftmargin=*, itemsep=0pt, topsep=0pt, parsep=0pt]
    \item $N = \{1,\dots,n\}$: the set of facilities / locations.
    \item $f_{ij}$: flow (interaction weight) between facility $i$ and facility $j$.
    \item $d_{kl}$: distance between location $k$ and location $l$.
    \item $x_{ik}$: binary variable that equals $1$ if and only if facility $i$ is assigned to location $k$.
\end{itemize}
Below is the standard Koopmans–Beckmann binary quadratic formulation:
\begin{align*}
\min \; & \sum_{i=1}^{n} \sum_{j=1}^{n} \sum_{k=1}^{n} \sum_{l=1}^{n} f_{ij}\, d_{kl}\, x_{ik}\, x_{jl} \\
\mathrm{s.t.} \; & \sum_{k=1}^{n} x_{ik} = 1, \quad \forall i \in N, \\
& \sum_{i=1}^{n} x_{ik} = 1, \quad \forall k \in N, \\
& x_{ik} \in \{0, 1\}, \quad \forall i,k \in N.
\end{align*}

\subsubsection*{Baseline formulations}
\begin{itemize}[leftmargin=*, itemsep=2pt, topsep=0pt, parsep=0pt]
    \item K-B Quad.: The standard Koopmans–Beckmann binary quadratic formulation, as described above.
    \item McC. Lin.: A classical linearization that introduces auxiliary variables $y_{ijkl}$ to represent products $x_{ik}x_{jl}$ and enforces equivalence via standard McCormick envelope inequalities.
    \item AJ Lin.: The Adams–Johnson linearization that lifts continuous $y_{ijkl}$ variables with tight linking constraints, yielding a stronger relaxation.
    \item XY-KB Lin.: The Xia–Yuan linearization that precomputes minimum-assignment lower and upper bounds, and then applies auxiliary variables with tightening lower-bound constraints.
    \item IPQAPR: A compact linearization that reduces the dimension of $y_{ijkl}$ by exploiting symmetry and enforces cross-block consistency constraints.
    \item StepORLM: Generated by StepORLM, same as McC.
    \item EvoCut: Same as K-B formulation, since the Birkhoff polytope already represents the convex hull of the assignment constraints.
\end{itemize}

\subsubsection*{FormuEvo formulation}
\paragraph{Notation}
\begin{itemize}[leftmargin=*, itemsep=0pt, topsep=0pt, parsep=0pt]
    \item $z_{ij}^{kl}$: lifted continuous variable in $[0,1]$, representing $x_{ik}x_{jl}$ for $i<j$ and $k\neq l$.
    \item $\mathcal{G}_F$: set of unordered facility pairs $(i,j)$ with $i<j$ whose flow rows and columns are identical, i.e., $f_{i,\cdot}=f_{j,\cdot}$ and $f_{\cdot,i}=f_{\cdot,j}$.
    \item $\mathcal{G}_D$: set of unordered location pairs $(k,l)$ with $k<l$ whose distance rows and columns are identical, i.e., $d_{k,\cdot}=d_{l,\cdot}$ and $d_{\cdot,k}=d_{\cdot,l}$.
\end{itemize}
FormuEvo uses a transportation-style linearization that lifts each unordered facility pair $(i,j)$ into a separate $z$-block. Cross-block consistency is enforced by anchor-slice coupling, and symmetry is broken via cumulative prefix inequalities.
\begin{align*}
\min \; & \sum_{i=1}^{n} \sum_{k=1}^{n} f_{ii} d_{kk}\, x_{ik} + \\
& \sum_{i=1}^{n-1} \sum_{j=i+1}^{n} \sum_{k=1}^{n} \sum_{l=1,\,l \neq k}^{n} \bigl(f_{ij} d_{kl} + f_{ji} d_{lk}\bigr)\, z_{ij}^{kl} \\
\mathrm{s.t.} \; & \sum_{k=1}^{n} x_{ik} = 1, \quad \forall i \in N, \\
& \sum_{i=1}^{n} x_{ik} = 1, \quad \forall k \in N, \\
& \sum_{l=1,\, l \neq k}^{n} z_{ij}^{kl} = x_{ik}, \quad \forall i<j,\ k \in N, \\
& \sum_{k=1,\, k \neq l}^{n} z_{ij}^{kl} = x_{jl}, \quad \forall i<j,\ l \in N, \\
& \sum_{j=i+1}^{n} z_{ij}^{kl} + \sum_{j=1}^{i-1} z_{ji}^{lk} = x_{ik}, \\
& \hspace{2.6cm} \forall i \in N,\ k,l \in N,\ k \neq l, \\
& \sum_{r=1}^{t} x_{ir} \ge \sum_{r=1}^{t} x_{jr}, \\
& \hspace{2.1cm} \forall \{i,j\} \in \mathcal{G}_F,\ i < j,\ t \in N, \\
& \sum_{r=1}^{t} x_{rk} \ge \sum_{r=1}^{t} x_{rl}, \\
& \hspace{2.1cm} \forall \{k,l\} \in \mathcal{G}_D,\ k < l,\ t \in N, \\
& x_{ik} \in \{0, 1\},\quad \forall i,k \in N, \\
& 0 \le z_{ij}^{kl} \le 1,\quad \forall i<j,\ k,l \in N,\ k \neq l.
\end{align*}

\subsection{Neural Network Verification (NNV)}

\subsubsection*{Problem description}

The neural network verification problem aims to determine the minimum adversarial distortion required to induce a targeted misclassification in a pre-trained multi-layer perceptron (MLP) classifier. Given a correctly classified input sample $x_0 \in [0,1]^{n_0}$ with true label $\lambda$, the goal is to find a perturbed input $x \in [0,1]^{n_0}$ that causes the network to predict a specified wrong label $\mu \neq \lambda$, while minimizing the $\ell_1$ distance between $x$ and $x_0$. The network consists of $L$ layers, where each hidden layer $k \in \{1,\dots, L-1\}$ applies a linear transformation followed by a ReLU activation, and the output layer $L$ applies a linear transformation only. The adversarial condition requires that the output logit of $\mu$ is at least that of $\lambda$. The objective is to minimize the $\ell_1$ distortion $\sum_i |x_i - x_{0,i}|$.

\subsubsection*{Standard MIP formulation}
\paragraph{Notation}
\begin{itemize}[leftmargin=*, itemsep=0pt, topsep=0pt, parsep=0pt]
    \item $n_0$: dimension of the input space.
    \item $L$: total number of (hidden + output) layers.
    \item $n_k$: number of neurons in layer $k$ ($k = 1,\dots,L$).
    \item $W^{(k)} \in \mathbb{R}^{n_k \times n_{k-1}}$: weight matrix of layer $k$.
    \item $b^{(k)} \in \mathbb{R}^{n_k}$: bias vector of layer $k$.
    \item $x_0 \in [0,1]^{n_0}$: original input, classified as $\lambda$.
    \item $\mu \neq \lambda$: target adversarial label.
    \item $x_i$: perturbed input component, $x_i \in [0,1]$.
    \item $d_i$: auxiliary variable for linearizing $|x_i - x_{0,i}|$.
    \item $\hat{z}^{(k)}_j$: pre-activation of neuron $j$ in layer $k$.
    \item $z^{(k)}_j$: post-activation of neuron $j$ in layer $k$.
    \item $a^{(k)}_j \in \{0,1\}$: ReLU indicator for neuron $j$ in layer $k$; $a=0$ encodes the active branch ($\hat{z}\ge0$), $a=1$ encodes the inactive branch ($\hat{z}\le0$).
    \item $M^{(k)}_j$: per-neuron big-$M$ bound obtained via interval arithmetic over $x \in [0,1]^{n_0}$.
\end{itemize}
\begin{align*}
\min \; & \sum_{i=1}^{n_0} d_i \\
\mathrm{s.t.} \; & d_i \ge x_i - x_{0,i},\quad d_i \ge x_{0,i} - x_i, \\
& \hspace{4.6cm} \forall i = 1,\dots,n_0,\\
& \hat{z}^{(k)}_j = \sum_{i=1}^{n_{k-1}} W^{(k)}_{ji} z^{(k-1)}_i + b^{(k)}_j,\\
& \hspace{2.3cm} \forall k=1,\dots,L,\ j=1,\dots,n_k, \\
& z^{(k)}_j \le \hat{z}^{(k)}_j + M^{(k)}_j a^{(k)}_j,\\
& \hspace{1.6cm} \forall k=1,\dots,L-1,\ j=1,\dots,n_k, \\
& z^{(k)}_j \ge \hat{z}^{(k)}_j, \\
& \hspace{1.6cm} \forall k=1,\dots,L-1,\ j=1,\dots,n_k, \\
& z^{(k)}_j \le M^{(k)}_j (1 - a^{(k)}_j),\\
& \hspace{1.6cm} \forall k=1,\dots,L-1,\ j=1,\dots,n_k, \\
& z^{(k)}_j \ge 0,\; \forall k=1,\dots,L-1,\ j=1,\dots,n_k, \\
& \hat{z}^{(L)}_\mu \ge \hat{z}^{(L)}_\lambda, \\
& 0 \le x_i \le 1, \quad \forall i = 1,\dots,n_0, \\
& a^{(k)}_j \in \{0, 1\},\\
& \hspace{1.6cm} \forall k=1,\dots,L-1,\ j=1,\dots,n_k,
\end{align*}
where $z^{(0)} = x$ (i.e., $z^{(0)}_i = x_i$), $z^{(L)} = \hat{z}^{(L)}$ (the output layer has no ReLU).

\subsubsection*{Baseline formulations}
\begin{itemize}[leftmargin=*, itemsep=2pt, topsep=0pt, parsep=0pt]
    \item Std. Formu.: The standard formulation as above.
    \item EvoCut: A strengthened variant that adds triangle relaxation inequalities, tightening the LP relaxation over the big-$M$ envelope.
\end{itemize}

\subsubsection*{FormuEvo formulation}
\paragraph{Notation}
\begin{itemize}[leftmargin=*, itemsep=0pt, topsep=0pt, parsep=0pt]
    \item $d_i^+,\ d_i^-$: split $\ell_1$ perturbation components, with $d_i^+ \in [0,\,1-x_{0,i}]$, $d_i^- \in [0,\,x_{0,i}]$; the perturbed input is $x_i = x_{0,i} + d_i^+ - d_i^-$ and the distortion is $d_i^+ + d_i^- = |x_i - x_{0,i}|$.
    \item $l^{(k)}_j$: pre-computed lower bounds on $\hat{z}^{(k)}_j$.
    \item $u^{(k)}_j$: pre-computed upper bounds on $\hat{z}^{(k)}_j$.
\end{itemize}
The formulation uses the split $\ell_1$ variables $d_i^+,d_i^-$ and per-neuron bounds $l^{(k)}_j,u^{(k)}_j$.
\begin{align*}
\min \; & \sum_{i=1}^{n_0} \bigl(d_i^+ + d_i^-\bigr) \\
\mathrm{s.t.} \; & 0 \le d_i^+ \le 1 - x_{0,i}, \, 0 \le d_i^- \le x_{0,i},\\
& \hspace{4.6cm} \forall i = 1,\dots,n_0, \\
& z^{(0)}_i = x_{0,i} + d_i^+ - d_i^-, \quad \forall i = 1,\dots,n_0, \\
& \hat{z}^{(k)}_j = \sum_{i=1}^{n_{k-1}} W^{(k)}_{ji} z^{(k-1)}_i + b^{(k)}_j,\\
& \hspace{2.3cm} \forall k=1,\dots,L,\ j=1,\dots,n_k, \\
& z^{(k)}_j \ge \hat{z}^{(k)}_j,\\
& \hspace{1.6cm} \forall k=1,\dots,L-1,\ j=1,\dots,n_k,\\
& z^{(k)}_j \le \hat{z}^{(k)}_j - l^{(k)}_j a^{(k)}_j,\\
& \hspace{1.6cm} \forall k=1,\dots,L-1,\ j=1,\dots,n_k, \\
& z^{(k)}_j \le u^{(k)}_j (1 - a^{(k)}_j),\\
& \hspace{1.6cm} \forall k=1,\dots,L-1,\ j=1,\dots,n_k, \\
& z^{(k)}_j \ge 0,\; \forall k=1,\dots,L-1,\ j=1,\dots,n_k,\\
& \hat{z}^{(L)}_\mu \ge \hat{z}^{(L)}_\lambda, \\
& d_i^+,\ d_i^- \ge 0, \quad \forall i = 1,\dots,n_0, \\
& a^{(k)}_j \in \{0, 1\},\\
& \hspace{1.6cm} \forall k=1,\dots,L-1,\ j=1,\dots,n_k.\\
\end{align*}

\subsection{IMO 2025 Problem 6 (IMO)}
\subsubsection*{Problem description}
The IMO 2025 Problem 6 asks to cover an $n \times n$ grid of unit squares using non-overlapping axis-aligned rectangular tiles, such that exactly one unit square remains uncovered (a ``hole'') in each row and each column. This problem can be modeled as an MIP. The objective is to minimize the total number of rectangles used. Each rectangle spans a contiguous interval of columns over a contiguous interval of rows. The decision involves selecting the hole positions and determining how rectangles are placed so that every non-hole cell is covered exactly once. The formulation additionally tracks the vertical continuation of horizontal intervals across adjacent rows, allowing rectangles to be represented compactly through interval propagation and flow-balance relations.

\subsubsection*{Standard MIP formulation}
\paragraph{Notation}
\begin{itemize}[leftmargin=*, itemsep=0pt, topsep=0pt, parsep=0pt]
    \item $n$: grid size ($n \times n$).
    \item $R = \{1,\dots,n\}$: the set of row indices.
    \item $C = \{1,\dots,n\}$: the set of column indices.
    \item $\mathcal{I} = \{(a,b) \mid 1 \le a \le b \le n\}$: the set of all contiguous column intervals.
    \item $h_{ij} \in \{0,1\}$: equals $1$ if and only if the hole in row $i$ is located at column $j$.
    \item $x_{iab} \in \{0,1\}$: equals $1$ if and only if interval $(a,b)$ is active on row $i$, i.e., columns $[a,\dots,b]$ are covered by the same rectangle on row $i$.
    \item $s_{iab} \in \{0,1\}$: equals $1$ if and only if a rectangle with horizontal span $(a,b)$ starts at row $i$.
    \item $t_{iab} \in \{0,1\}$: equals $1$ if and only if a rectangle with horizontal span $(a,b)$ ends at row $i$.
\end{itemize}
\begin{align*}
\min \; & \sum_{i=1}^{n} \sum_{(a,b) \in \mathcal{I}} s_{iab} \\
\mathrm{s.t.} \; & \sum_{j=1}^{n} h_{ij} = 1, \quad \forall i \in R, \\
& \sum_{i=1}^{n} h_{ij} = 1, \quad \forall j \in C, \\
& \hspace{-0.7cm} \sum_{(a,b) \in \mathcal{I}:\, a \le j \le b} x_{iab} + h_{ij} = 1,\quad \forall i \in R,\ j \in C, \\
& x_{1ab} - s_{1ab} = 0, \quad \forall (a,b) \in \mathcal{I}, \\
& x_{iab} - x_{i-1,ab} - s_{iab} + t_{i-1,ab} = 0, \\
& \hspace{2.7cm} \forall i = 2,\dots,n,\ (a,b) \in \mathcal{I}, \\
& x_{nab} - t_{nab} = 0, \quad \forall (a,b) \in \mathcal{I}, \\
& h_{ij} \in \{0, 1\},\quad \forall i \in R,\ j \in C, \\
& x_{iab},\, s_{iab},\, t_{iab} \in \{0, 1\},\, \forall i \in R,\ (a,b) \in \mathcal{I}.
\end{align*}

\subsubsection*{Baseline formulations}
\begin{itemize}[leftmargin=*, itemsep=2pt, topsep=0pt, parsep=0pt]
    \item Std. Formu.: The standard formulation as above.
    \item EvoCut: A variant with cuts enforcing mandatory new interval starts when hole movement disrupts coverage continuity across rows.
\end{itemize}

\subsubsection*{FormuEvo formulation}
\paragraph{Notation}
\begin{itemize}[leftmargin=*, itemsep=0pt, topsep=0pt, parsep=0pt]
    \item $\mathcal{R} = \{(i_1,i_2,a,b) \mid 1 \le i_1 \le i_2 \le n,\ (a,b) \in \mathcal{I}\}$: the set of all axis-aligned subrectangles.
    \item $y_{i_1 i_2 a b} \in \{0,1\}$: equals $1$ if and only if a rectangle spanning rows $i_1$ through $i_2$ and columns $a$ through $b$ is selected.
    \item $c_{ij} \in [0,1]$: variable representing the number of selected rectangles covering cell $(i,j)$.
    \item $\mathcal{R}_{ij}^{\mathrm{TL}} = \{(i_1,i_2,a,b) \in \mathcal{R} \mid i_1=i,\ a=j\}$: rectangles whose top-left corner is at $(i,j)$.
    \item $\mathcal{R}_{ij}^{\mathrm{TR}} = \{(i_1,i_2,a,b) \in \mathcal{R} \mid i_1=i,\ b=j-1\}$ ($j>1$): rectangles whose top row is $i$ and rightmost column is $j-1$.
    \item $\mathcal{R}_{ij}^{\mathrm{BL}} = \{(i_1,i_2,a,b) \in \mathcal{R} \mid i_2=i-1,\ a=j\}$ ($i>1$): rectangles whose bottom row is $i-1$ and leftmost column is $j$.
    \item $\mathcal{R}_{ij}^{\mathrm{BR}} = \{(i_1,i_2,a,b) \in \mathcal{R} \mid i_2=i-1,\ b=j-1\}$ ($i,j>1$): rectangles whose bottom-right corner is at $(i-1,j-1)$.
\end{itemize}
FormuEvo formulation directly models each possible axis-aligned rectangle as a binary variable, then reconstructs cell occupancy via a 2D difference array. Strong aggregated cuts enforce exact cover by projecting onto rows, columns, and the total area.
\begin{align*}
\min \; & \sum_{(i_1,i_2,a,b) \in \mathcal{R}} y_{i_1 i_2 a b} \\
\mathrm{s.t.} \; & \sum_{j=1}^{n} h_{ij} = 1, \quad \forall i \in R, \\
& \sum_{i=1}^{n} h_{ij} = 1, \quad \forall j \in C, \\
& c_{ij} = c_{i-1,j} + c_{i,j-1} - c_{i-1,j-1} \\
& \quad + \sum_{\mathcal{R}_{ij}^{\mathrm{TL}}} y_{i_1 i_2 a b}
         - \sum_{\mathcal{R}_{ij}^{\mathrm{TR}}} y_{i_1 i_2 a b} \\
& \quad - \sum_{\mathcal{R}_{ij}^{\mathrm{BL}}} y_{i_1 i_2 a b}
         + \sum_{\mathcal{R}_{ij}^{\mathrm{BR}}} y_{i_1 i_2 a b},\\
& \hspace{4.3cm} \forall i \in R,\ j \in C, \\
& c_{ij} + h_{ij} = 1, \quad \forall i \in R,\ j \in C, \\
& \hspace{-0.8cm} \sum_{(i_1,i_2,a,b) \in \mathcal{R} :\, i_1 \le i \le i_2} \hspace{-0.5cm} (b - a + 1)\, y_{i_1 i_2 a b}
    = n - 1,\\
& \hspace{5.5cm} \forall i \in R, \\
& \hspace{-0.7cm} \sum_{(i_1,i_2,a,b) \in \mathcal{R} :\, a \le j \le b} \hspace{-0.6cm} (i_2 - i_1 + 1)\, y_{i_1 i_2 a b}
    = n - 1,\\
& \hspace{5.5cm} \forall j \in C, \\
& \hspace{-0.2cm} \sum_{(i_1,i_2,a,b) \in \mathcal{R}} \hspace{-0.1cm} (i_2 - i_1 + 1)(b - a + 1)\, y_{i_1 i_2 a b} = \\
& \hspace{5.3cm} n(n - 1), \\
& y_{i_1 i_2 a b} \in \{0, 1\},\quad \forall (i_1,i_2,a,b) \in \mathcal{R}, \\
& c_{ij} \in \{0,1\},\quad \forall i \in R,\ j \in C.
\end{align*}
where $c_{0,j} = c_{i,0} = c_{0,0} = 0$, and all summations over $\mathcal{R}_{ij}^{\cdot}$ involving out-of-bounds indices are defined to be zero. The reconstruction of $c_{ij}$ follows the standard 2D difference-array scheme: each rectangle variable $y_{i_1 i_2 a b}$ induces a unit increment over its support via corner updates $+1$ at $(i_1,a)$, $-1$ at $(i_1,b+1)$, $-1$ at $(i_2+1,a)$, and $+1$ at $(i_2+1,b+1)$. Consequently, the resulting prefix-sum transformation yields $c_{ij}=1$ if and only if cell $(i,j)$ lies within exactly one selected rectangle.

\section{Implementation Details of FormuEvo} \label{app:implementation}
Algorithm~\ref{alg:algorithm} outlines the framework of FormuEvo. Below, we present a reference of our prompts used in FormuEvo. All complete prompts are available in our code repository at \url{https://github.com/Xyz-yuanhf/formuevo}.

\begin{algorithm*}[!t]
\caption{Pseudo-code of FormuEvo}\label{alg:algorithm}
\begin{algorithmic}[1]
\REQUIRE Problem description $p$, formulation template $f_0$, population size $N$, generations $T$, mutation rate $\rho$, memory usage rate $\gamma$, external knowledge base $\mathcal{B}_\text{init}$ (optional).
\ENSURE Final MIP formulation $\hat{f}^\star$, distilled knowledge base $\mathcal{B}$.
\STATE Initialize population $\mathcal{P}_0 \leftarrow \varnothing$ 
\STATE Initialize memory library $\mathcal{M} \leftarrow \mathcal{B}_\text{init}$ (if $\mathcal{B}_\text{init}$ is available, $\varnothing$ otherwise)

\STATE \textcolor{gray}{// \textit{Initialization: generate initial population}}
\WHILE{$|\mathcal{P}_0| < N$}
    \STATE $f_{\text{init}} \leftarrow \textsc{GeneratorLLM}(p,\, f_0)$ \hfill \textcolor{gray}{// \textit{Initial formulation}}
    \STATE $\mathcal{P}_0 \leftarrow \mathcal{P}_0 \cup \{\textsc{EvaluateAndRepair}(f_{\text{init}})\}$ \hfill \textcolor{gray}{// \textit{Evaluate and repair}}
\ENDWHILE
\STATE \textcolor{gray}{// \textit{Update best formulation}}
\STATE $\hat{f}^\star \leftarrow \arg\min_{f \in \mathcal{P}_0} \phi(f)$

\STATE \textcolor{gray}{// \textit{Evolution: repeat for $T$ generations}}
\FOR{$t = 1$ \textbf{to} $T$}
    \STATE $\mathcal{O} \leftarrow \varnothing$ \hfill \textcolor{gray}{// \textit{Offspring pool}}
    \WHILE{$|\mathcal{O}| < N$}
        \STATE \textcolor{gray}{// \textit{Crossover operation}}
        \STATE Sample two parents $f_i,\, f_j$ from $\mathcal{P}_{t-1}$
        \STATE \textcolor{gray}{// \textit{Diagnosis of parents}}
        \IF{$u < \gamma$ where $u \sim U(0, 1)$}
            \STATE $d_{ij} \leftarrow \textsc{DiagnosticLLM}(f_i,\, \text{stats}(f_i),\, f_j,\, \text{stats}(f_j),\, \mathcal{M})$ \hfill \textcolor{gray}{// \textit{Diagnosis of parents}}
        \ELSE
            \STATE $d_{ij} \leftarrow \textsc{DiagnosticLLM}(f_i,\, \text{stats}(f_i),\, f_j,\, \text{stats}(f_j))$ \hfill \textcolor{gray}{// \textit{Without memory}}
        \ENDIF
        \STATE $f_o \leftarrow \textsc{GeneratorLLM}(p,\, f_i,\, f_j,\, d_{ij})$ \hfill \textcolor{gray}{// \textit{Offspring by crossover}}
        \STATE $\mathcal{O} \leftarrow \mathcal{O} \cup \{\textsc{EvaluateAndRepair}(f_o)\}$ \hfill \textcolor{gray}{// \textit{Evaluate and repair}}
        \STATE $\mathcal{M} \leftarrow \textsc{ReflectorLLM}(f_i,\, \text{stats}(f_i),\, f_j,\, \text{stats}(f_j),\,f_o,\, \text{stats}(f_o),\, \mathcal{M})$ \hfill \textcolor{gray}{// \textit{Update memory}}
        \STATE \textcolor{gray}{// \textit{Mutation operation}}
        \IF{$|\mathcal{O}| < N$ \AND $u < \rho$ where $u \sim U(0, 1)$}
            \STATE Sample elite $f_e$ from $\mathcal{P}_{t-1}$
            \IF{$u < \gamma$ where $u \sim U(0, 1)$}
                \STATE $d_{e} \leftarrow \textsc{DiagnosticLLM}(f_e,\, \text{stats}(f_e),\, \mathcal{M})$ \hfill \textcolor{gray}{// \textit{Diagnosis of elite}}
            \ELSE
                \STATE $d_{e} \leftarrow \textsc{DiagnosticLLM}(f_e,\, \text{stats}(f_e))$ \hfill \textcolor{gray}{// \textit{Without memory}}
            \ENDIF
            \STATE $f_o \leftarrow \textsc{GeneratorLLM}(p,\, f_e,\, d_{e})$ \hfill \textcolor{gray}{// \textit{Offspring by mutation}}
            \STATE $\mathcal{O} \leftarrow \mathcal{O} \cup \{\textsc{EvaluateAndRepair}(f_o)\}$ \hfill \textcolor{gray}{// \textit{Evaluate and repair}}
            \STATE $\mathcal{M} \leftarrow \textsc{ReflectorLLM}(f_e,\, \text{stats}(f_e),\, f_o,\, \text{stats}(f_o),\, \mathcal{M})$ \hfill \textcolor{gray}{// \textit{Update memory}}
        \ENDIF
    \ENDWHILE
    \STATE \textcolor{gray}{// \textit{Select top-$N$ individuals for next generation}}
    \STATE $\mathcal{P}_t \leftarrow \textsc{TopN}(\mathcal{P}_{t-1} \cup \mathcal{O},\, N)$
    \STATE \textcolor{gray}{// \textit{Update best formulation}}
    \STATE $\hat{f}^\star \leftarrow \arg\min(\phi(\hat{f}^\star),\, \min_{f \in \mathcal{O}} \phi(f))$
\ENDFOR

\STATE \textcolor{gray}{// \textit{Knowledge distillation for transfer}}
\STATE $\mathcal{B} \leftarrow \textsc{DistillerLLM}(\mathcal{M})$ \hfill \textcolor{gray}{// \textit{Distill knowledge from memory}}
\STATE \textbf{return} $\hat{f}^\star,\, \mathcal{B}$
\end{algorithmic}
\vspace{5pt}
\end{algorithm*}

\subsection*{Generator LLM for Initialization}
For initialization, we use the prompt for the generator LLM to produce the initial population.
\paragraph*{\,}
\begin{tcolorbox}[colback=gray!5!white, colframe=gray!40!black, title=System Prompt, fonttitle=\bfseries, fontupper=\small, arc=2mm, boxrule=0.8pt, left=6pt, right=6pt, top=2pt, bottom=2pt, breakable]
You are an expert in mathematical optimization, mixed-integer programming (MIP), and Gurobi modeling. You specialize in translating the problem description and its preliminary Gurobi models into mathematically correct, high-efficiency, and numerically stable Gurobi models implemented in Python using `gurobipy'.

The user will provide:
\begin{itemize}[leftmargin=*, itemsep=0pt, topsep=0pt, parsep=0pt]
\item Problem description: The real-world logic and constraints of the optimization problem to be modeled.
\item Python template: A Python `gurobipy' function template. You must preserve the function name and input arguments exactly. You may refactor the internal implementation freely.
\item Memory library (if available): A collection of experience triplets (condition → strategy → effect) distilled from past evolution experiments.
\end{itemize}

Your task is:
\begin{itemize}[leftmargin=*, itemsep=0pt, topsep=0pt, parsep=0pt]
\item Generate a better (i.e., mathematically correct, high-efficiency, and numerically stable) Gurobi model than the preliminary one.
\item Your generated model should be implemented in Python `gurobipy' and follow the standard Gurobi modeling practices. Ensure all variables, constraints, and objective functions are well-defined.
\item Your generated model must be mathematically correct and designed to solve as efficiently as possible under Gurobi's default parameters (do not rely on parameter tuning).
\item To achieve this you should apply advanced modeling techniques where appropriate, including but not limited to \{example\_strengthening\_techniques\}.
\item You should produce clean, well-structured, and executable Python code using `gurobipy'. Keep code concise but readable.
\item You must provide a concise explanation of the optimization ideas and key improvements in the `idea' section. Focus on your enhancement motivation and why the new formulation is computationally superior.
\end{itemize}
Critical constraints:
\begin{itemize}[leftmargin=*, itemsep=0pt, topsep=0pt, parsep=0pt]
\item Do not change the function signature.
\item Do not rely on solver parameter tuning. However, setting parameters strictly required for the model to be accepted by Gurobi is allowed if your formulation necessitates it.
\item Do not assume missing data unless clearly implied; make minimal necessary assumptions.
\item If ambiguity exists, choose the most standard and computationally efficient formulation.
\item Calling `model.optimize()' or any other solving/execution functions is strictly forbidden within the function. The function must return the model object before any optimization is performed.
\item Do not use Gurobi callbacks in any form. If your formulation would naturally rely on callbacks, you must replace it with a static, callback-free alternative.
\end{itemize}
Response format:
\begin{itemize}[leftmargin=*, itemsep=0pt, topsep=0pt, parsep=0pt]
\item Return only a valid JSON object in the exact structure below (\{detailed\_format\_instruction\}):
\end{itemize}

\texttt{\{\\
``code'': ``\`{}\`{}\`{}python\textbackslash n<A Python (gurobipy) function implementing the improved Gurobi model>\textbackslash n\`{}\`{}\`{}'',\\
``idea'': ``A concise but technically clear idea of key modeling improvements and any non-trivial reformulations''\\
\}}
\end{tcolorbox}
\paragraph*{\,}
\begin{tcolorbox}[colback=blue!3!white, colframe=blue!40!black, title=User Prompt, fonttitle=\bfseries, fontupper=\small, arc=2mm, boxrule=0.8pt, left=6pt, right=6pt, top=2pt, bottom=2pt, breakable]
Problem description:\\
\{problem\_description\}\\
\\
Python template:\\
\{python\_template\}\\
\\
Memory library (if available):\\
\{memory\_library\}
\end{tcolorbox}

\subsection*{Diagnostic LLM}
We present the prompt of the diagnostic LLM for mutation. The prompt for crossover follows the same structure but operates on a formulation pair.
\paragraph*{\,}
\begin{tcolorbox}[colback=gray!5!white, colframe=gray!40!black, title=System Prompt, fonttitle=\bfseries, fontupper=\small, arc=2mm, boxrule=0.8pt, left=6pt, right=6pt, top=2pt, bottom=2pt, breakable]
You are an expert in mathematical optimization, mixed-integer programming (MIP), Gurobi modeling, and structural evolution. Your role is to examine the solver performance profile of a given formulation and produce a concise, actionable diagnostic report that will guide a downstream mutation agent in evolving this formulation into a computationally superior variant.

The user will provide:
\begin{itemize}[leftmargin=*, itemsep=0pt, topsep=0pt, parsep=0pt]
\item Problem description: The real-world logic and constraints of the optimization problem to be modeled.
\item Current formulation data: You will receive its Python `gurobipy' code and the mathematical idea behind it.
\item Solver performance profile, including:\\
num\_vars: Number of variables in the model.\\
num\_constrs: Number of constraints in the model.\\
root\_bound: Average relaxation bound at root node.\\
root\_gap: Average root relaxation gap (\%) = |opt\_obj - root\_bound| / |opt\_obj| * 100.\\
node\_count: Average number of Branch-and-Bound nodes explored.\\
presolve\_rows\_removed: Average number of rows (constraints) removed by Gurobi presolve.\\
presolve\_cols\_removed: Average number of columns (variables) removed by Gurobi presolve.\\
presolve\_bounds\_changed: Average number of variable bounds changed by Gurobi presolve.
\item Memory library (if available): A collection of experience triplets (condition → strategy → effect) distilled from past evolution experiments.
\end{itemize}
Your task is:
\begin{itemize}[leftmargin=*, itemsep=0pt, topsep=0pt, parsep=0pt]
\item Perform a bottleneck identification: Determine the primary computational bottleneck of the current formulation. Use, but not limited to, the following diagnostic logic: \{example\_diagnostic\_logic\}
\item Identify secondary bottlenecks (if possible): After identifying the primary bottleneck, note any secondary issues that could become the new bottleneck once the primary is resolved. This helps mutation avoid shifting the problem without actually improving it.
\item Report a concise mutation guidance: Based on your diagnosis, produce specific, actionable recommendations for mutation. Your recommendations should:\\
State the single most impactful mutation direction.\\
Provide 1-3 specific MIP modeling techniques to apply, ranked by expected impact. Indicate what the mutation should preserve from the current formulation. Suggest the expected trade-off. Warn against the potential risks of the recommendation. You can retrieve the memory library for entries whose condition matches the diagnosed conditions.
\end{itemize}
Output format:\\
Return a structured diagnostic report in the format:\\
Formulation profile:
\begin{itemize}[leftmargin=*, itemsep=0pt, topsep=0pt, parsep=0pt]
\item Size: [num\_vars] variables, [num\_constrs] constraints
\item Relaxation: root\_gap = [value]\%
\item Branching: node\_count = [value]
\item Presolve: [rows\_removed] rows removed, [cols\_removed] cols removed, [bounds\_changed] bounds changed
\item Overall assessment: [one-sentence summary of where time is being spent]
\end{itemize}
Primary bottleneck:
\begin{itemize}[leftmargin=*, itemsep=0pt, topsep=0pt, parsep=0pt]
\item Category: [weak relaxation / excessive branching / per-node cost / loose bounds / model size]
\item Evidence: [cite specific metric values]
\end{itemize}
Secondary bottleneck(s):
\begin{itemize}[leftmargin=*, itemsep=0pt, topsep=0pt, parsep=0pt]
\item Category: [description, or "none identified."]
\end{itemize}
Mutation recommendations (priority-ranked):
\begin{itemize}[leftmargin=*, itemsep=0pt, topsep=0pt, parsep=0pt]
\item The most impactful technique with specific detail
\item Second technique
\item Third technique, if applicable
\end{itemize}
Preserve (do not degrade):
\begin{itemize}[leftmargin=*, itemsep=0pt, topsep=0pt, parsep=0pt]
\item Aspects of current formulation that are strong and must be maintained
\end{itemize}
Expected trade-offs:
\begin{itemize}[leftmargin=*, itemsep=0pt, topsep=0pt, parsep=0pt]
\item What metrics should improve and what might worsen
\end{itemize}
Risk warnings:
\begin{itemize}[leftmargin=*, itemsep=0pt, topsep=0pt, parsep=0pt]
\item Potential pitfalls of the mutation, or "none identified."
\end{itemize}
\end{tcolorbox}
\paragraph*{\,}
\begin{tcolorbox}[colback=blue!3!white, colframe=blue!40!black, title=User Prompt, fonttitle=\bfseries, fontupper=\small, arc=2mm, boxrule=0.8pt, left=6pt, right=6pt, top=2pt, bottom=2pt, breakable]
Problem description:\\
\{problem\_description\}\\
\\
Current formulation:\\
\{current\_formulation\_data\}\\
\\
Runtime:\\
\{runtime\_profile\}\\
\\
Solver profile:\\
\{solver\_profile\}\\
\\
Memory library (if available):\\
\{memory\_library\}
\end{tcolorbox}

\subsection*{Crossover and Mutation}
We employ four types of crossover operators: complement, hybrid, intersection, and min-violation, and four types of mutation operators: general, exploratory, searching, and polyhedral, representing different interaction relationships and evolutionary focuses. Here we present the prompt of general mutation, and other operators are similar in structure, available in our code repository.
\paragraph*{\,}
\begin{tcolorbox}[colback=gray!5!white, colframe=gray!40!black, title=System Prompt, fonttitle=\bfseries, fontupper=\small, arc=2mm, boxrule=0.8pt, left=6pt, right=6pt, top=2pt, bottom=2pt, breakable]
You are an expert in mathematical optimization, mixed-integer programming (MIP), and Gurobi modeling. You specialize in evolving a baseline formulation through rigorous mathematical refinement, polyhedral lifting, and innovative paradigm shifts, transforming it into a mathematically correct and substantially more efficient model implemented in Python using `gurobipy'.

The user will provide:
\begin{itemize}[leftmargin=*, itemsep=0pt, topsep=0pt, parsep=0pt]
\item Problem description: The real-world logic and constraints of the optimization problem to be modeled.
\item Python template: A Python `gurobipy' function template. You must preserve the function name and input arguments exactly. You may refactor the internal implementation freely.
\item Current formulation data: You will receive its Python `gurobipy' code, the mathematical idea behind it, and its evaluation fitness, where a lower fitness indicates a computationally superior model, e.g., lower average runtime.
\item Diagnostic report (if available): A structured analysis of the current model's solver performance profile, identifying the computational bottlenecks and priority mutation directions.
\end{itemize}

Your task is:
\begin{itemize}[leftmargin=*, itemsep=0pt, topsep=0pt, parsep=0pt]
\item Perform a general mutation. You must analyze the provided model and propose a structurally refined version of it.
\item Retain the core mathematical paradigm of the original model, but aggressively seek local improvements to optimize its computational efficiency.
\item Identify and eliminate redundant constraints, unnecessary variables, or slight symmetries.
\item Improve the structural implementation. For instance, tighten variable bounds, compute tighter big-$M$ coefficients based on problem data, or rewrite logical conditions using more efficient `gurobipy' features.
\item Your mutated model should act as an optimized, polished evolution of its predecessor, strictly maintaining mathematical correctness while accelerating the branch-and-bound process.
\item To achieve this you should apply advanced modeling techniques where appropriate, including but not limited to \{example\_strengthening\_techniques\}.
\item You should produce clean, well-structured, and executable Python code using `gurobipy'. Keep code concise but readable.
\item You must provide a concise explanation of the optimization ideas and key improvements in the `idea' section. Focus on your enhancement motivation and why the new formulation is computationally superior.
\end{itemize}
Critical constraints:
\begin{itemize}[leftmargin=*, itemsep=0pt, topsep=0pt, parsep=0pt]
\item Do not change the function signature.
\item Do not rely on solver parameter tuning. However, setting parameters strictly required for the model to be accepted by Gurobi is allowed if your formulation necessitates it.
\item Do not assume missing data unless clearly implied; make minimal necessary assumptions.
\item If ambiguity exists, choose the most standard and computationally efficient formulation.
\item Calling `model.optimize()' or any other solving/execution functions is strictly forbidden within the function. The function must return the model object before any optimization is performed.
\item Do not use Gurobi callbacks in any form. If your formulation would naturally rely on callbacks, you must replace it with a static, callback-free alternative.
\end{itemize}
Response format:
\begin{itemize}[leftmargin=*, itemsep=0pt, topsep=0pt, parsep=0pt]
\item Return only a valid JSON object in the exact structure below (\{detailed\_format\_instruction\}):
\end{itemize}

\texttt{\{\\
``code'': ``\`{}\`{}\`{}python\textbackslash n<A Python (gurobipy) function implementing the improved Gurobi model>\textbackslash n\`{}\`{}\`{}'',\\
``idea'': ``Explain the specific local structural improvements and why they reduce solving time''\\
\}}
\end{tcolorbox}
\paragraph*{\,}
\begin{tcolorbox}[colback=blue!3!white, colframe=blue!40!black, title=User Prompt, fonttitle=\bfseries, fontupper=\small, arc=2mm, boxrule=0.8pt, left=6pt, right=6pt, top=2pt, bottom=2pt, breakable]
Problem description:\\
\{problem\_description\}\\
\\
Python template:\\
\{python\_template\}\\
\\
Current formulation:\\
\{current\_formulation\_data\}\\
\\
Runtime:\\
\{runtime\_profile\}\\
\\
Solver profile:\\
\{solver\_profile\}\\
\\
Diagnostic report (if available):\\
\{diagnostic\_report\}
\end{tcolorbox}

\subsection*{Memory Reflection}
\paragraph*{\,}
\begin{tcolorbox}[colback=gray!5!white, colframe=gray!40!black, title=System Prompt, fonttitle=\bfseries, fontupper=\small, arc=2mm, boxrule=0.8pt, left=6pt, right=6pt, top=2pt, bottom=2pt, breakable]
You are an expert in mathematical optimization, mixed-integer programming (MIP), Gurobi modeling, and algorithmic analysis. Your role is to distill actionable, transferable insights and knowledge from MIP formulation evolutions by analyzing parent-offspring relationships and extracting concise experience triplets.

The user will provide:
\begin{itemize}[leftmargin=*, itemsep=0pt, topsep=0pt, parsep=0pt]
\item Problem description: The real-world logic and constraints of the optimization problem to be modeled.
\item Parent formulation data: For each parent formulation, you will receive its Python `gurobipy' code, the mathematical idea behind it, its evaluation fitness (i.e., average runtime), and its solver performance profiles.
\item Diagnostic report: the diagnostic analysis that guided this evolution step.
\item Offspring formulation data: Python `gurobipy' code and the mathematical idea of the generated offspring formulation.
\item Evolution outcome: Fitness (average runtime) and solver performance profiles for the generated offspring formulation, in the same format as the parent(s).
\end{itemize}
Your task is:
Extract exactly experience triplets from this evolution step. Each triplet captures a reusable lesson in the form:
\begin{itemize}[leftmargin=*, itemsep=0pt, topsep=0pt, parsep=0pt]
\item Condition: The observable situation before the modification, described in terms of solver profile symptoms and/or structural problem features. Conditions must be generalizable. Do not reference specific problem names, variable names, or instance sizes. Use solver metric patterns and structural descriptors only.
\item Strategy: The specific modeling action taken, described as a concrete, reproducible technique. Strategies must be precise enough that another modeler could apply them to a different problem exhibiting the same Condition.
\item Effect: The outcome and mechanistic explanation, state whether performance improved or degraded, cite the key metric changes, and briefly explain why.
\end{itemize}
Guidelines for high-quality triplets:
\begin{itemize}[leftmargin=*, itemsep=0pt, topsep=0pt, parsep=0pt]
\item Be concise: Each field should be 1-2 sentences maximum. Avoid verbose explanations.
\item Be generalizable: Triplets should apply beyond the specific problem instance. Abstract away problem-specific details into structural patterns.
\item Capture both successes and failures: Failed strategies (offspring worse than parent) are equally valuable, as long as they prevent repeating mistakes.
\item Be mechanistic: The Effect field must explain the solver mechanism, not just report metric changes.
\item One lesson per triplet: If the evolution involved multiple simultaneous changes, decompose into separate triplets where possible, attributing each metric change to its most likely cause.
\item Mark outcome clearly: Start the Effect field with [+] for improvements or [-] for degradations.
\end{itemize}
Response format:
\begin{itemize}[leftmargin=*, itemsep=0pt, topsep=0pt, parsep=0pt]
\item Your response must consist of ONLY the triplets in the exact format below. No preamble, no greeting, no commentary.
\item Return 1-3 triplets. Use exactly the format shown. Number each triplet.
\end{itemize}

\texttt{\\
- Condition: [generalizable situation descriptor]\\
- Strategy: [concrete modeling action]\\
- Effect: [+/-] [metric changes and mechanistic explanation]\\
}
\end{tcolorbox}

\paragraph*{\,}
\begin{tcolorbox}[colback=blue!3!white, colframe=blue!40!black, title=User Prompt, fonttitle=\bfseries, fontupper=\small, arc=2mm, boxrule=0.8pt, left=6pt, right=6pt, top=2pt, bottom=2pt, breakable]
Problem description:\\
\{problem\_description\}\\
\\
Parent formulation:\\
\{parent\_formulation\_data\}\\
\\
Parent runtime and solver profile:\\
\{parent\_runtime\_solver\_profile\}\\
\\
Diagnostic report:\\
\{diagnostic\_report\}\\
\\
Offspring formulation:\\
\{offspring\_formulation\_data\}\\
\\
Offspring runtime and solver profile:\\
\{offspring\_runtime\_solver\_profile\}
\end{tcolorbox}

\subsection*{Knowledge Distillation}
\paragraph*{\,}
\begin{tcolorbox}[colback=gray!5!white, colframe=gray!40!black, title=System Prompt, fonttitle=\bfseries, fontupper=\small, arc=2mm, boxrule=0.8pt, left=6pt, right=6pt, top=2pt, bottom=2pt, breakable]
You are an expert in mathematical optimization, mixed-integer programming (MIP), Gurobi modeling, and algorithmic analysis. Your role is to distill multiple memory libraries (may be problem-specific) into a single, unified, general-purpose knowledge base of MIP formulation principles that are transferable to any new optimization problem.

The user will provide:
Problem-specific memory libraries: Multiple memory libraries, each accumulated from evolutionary MIP formulation improvement on a specific problem type. Each library contains experience triplets with Condition, Strategy, and Effect. The libraries may come from diverse problem domains (e.g., scheduling, routing, packing, facility location, etc.)

Your task is to produce a general transferable knowledge base by performing the following:
\begin{itemize}[leftmargin=*, itemsep=2pt, topsep=2pt, parsep=0pt]
\item Cross-problem validation: Identify entries that appear (with consistent Effects) across multiple problem-specific libraries. These represent robust, problem-independent MIP modeling principles and should be prioritized and preserved with high fidelity.
\item Abstraction: For entries that reference residual problem-specific structure (even if already somewhat general), further abstract the Condition to the broadest valid scope.
\item Consolidation: Merge entries from different libraries that express the same underlying principle. Even if worded differently or applied in different problem contexts, unify them into a single canonical entry if the Condition-Strategy-Effect logic is equivalent. Combine metric evidence from multiple sources.
\item Conflict resolution: If the same Condition-Strategy pair shows positive Effect in some libraries but negative Effect in others, analyze whether the divergence can be explained by a distinguishable sub-condition (e.g., strategy works for sparse constraint matrices but fails for dense ones). If so, split into two entries with refined, non-overlapping Conditions.
\item Pruning: The final knowledge base should contain at most 30 entries. Remove entries that are too narrow to plausibly transfer or trivial or obvious.
\end{itemize}
Response format:
\begin{itemize}[leftmargin=*, itemsep=0pt, topsep=0pt, parsep=0pt]
\item Your response must consist of only the triplets in the exact format below. No preamble, no greeting, no commentary.
\item Number entries sequentially with IDs starting from 1.
\item Use exactly the format below for each entry.
\end{itemize}
\texttt{\\
\relax[KNOWLEDGE BASE START]\\
ID: 1\\
- Condition: [generalizable situation descriptor]\\
- Strategy: [concrete modeling action]\\
- Effect: [+/-] [metric changes and mechanistic explanation]\\
ID: 2\\
- Condition: [generalizable situation descriptor]\\
- Strategy: [concrete modeling action]\\
- Effect: [+/-] [metric changes and mechanistic explanation]\\
...\\
\relax[KNOWLEDGE BASE END]\\
}
\end{tcolorbox}
\paragraph*{\,}
\begin{tcolorbox}[colback=blue!3!white, colframe=blue!40!black, title=User Prompt, fonttitle=\bfseries, fontupper=\small, arc=2mm, boxrule=0.8pt, left=6pt, right=6pt, top=2pt, bottom=2pt, breakable]
Memory libraries:\\
\{memory\_libraries\}
\end{tcolorbox}

\section{Additional Results and Discussions}

\subsection{Statistical Significance Analysis} \label{app:significance}

We conduct paired Wilcoxon signed-rank tests on per-instance runtimes to compare FormuEvo with the best baselines over each problem and difficulty category (Table~\ref{tab:wilcoxon}). For most problems such as TSP, CFLP, and the NNV and IMO tasks, FormuEvo achieves statistically significant improvements ($p < 0.01$) across all categories. Notably, significance generally increases with instance difficulty, suggesting that the advantages of evolved formulations are more evident in challenging problem instances, where formulation strength critically impacts relaxation quality and branching behavior.

For BPP and QAP, the improvements on Easy instances are less significant. We attribute this to the ``floor effect'': modern MIP solvers can solve small instances within extremely short runtimes, and the performance is more sensitive to runtime variance across instances. As the problem scale and complexity grow (Medium and Hard), the advantages of FormuEvo become increasingly significant, demonstrating the strong scalability and upper-bound capabilities of FormuEvo, while further highlighting the importance of solver-efficient formulation for large-scale and more challenging optimization problems.

\begin{table}[!h]
\centering
\caption{Wilcoxon signed-rank test $p$-values.}\label{tab:wilcoxon}
\setlength{\tabcolsep}{6pt}
\resizebox{0.95\linewidth}{!}{
\begin{tabular}{c|c|c|c}
\toprule
Problem & Easy & Medium & Hard \\
\midrule
TSP & 1.88E-17 & 2.79E-13 & 4.13E-18 \\
JSSP & 7.83E-02 & 1.13E-03 & 2.44E-02 \\
BPP & 2.82E-01 & 9.29E-02 & 2.50E-02 \\
CFLP & 4.98E-13 & 3.03E-10 & 2.20E-08 \\
QAP & 4.44E-01 & 7.93E-02 & 3.05E-02 \\
\midrule
NNV & - & - & 1.42E-14 \\
IMO & - & - & 3.91E-03 \\
\bottomrule
\end{tabular}
}
\vspace{-5pt}
\end{table}

\begin{table*}[!t]
\centering
\caption{Performance based on COPT solver.}\label{tab:copt}
\setlength{\tabcolsep}{4pt}
\resizebox{0.9\linewidth}{!}{
\begin{tabular}{ll|cc|cc|ccc}
\toprule
&& \multicolumn{2}{c|}{Easy} & \multicolumn{2}{c|}{Medium} & \multicolumn{3}{c}{Hard} \\
\midrule
& Formulation & Time$\downarrow$ & Wins$\uparrow$ & Time$\downarrow$ & Wins$\uparrow$ & Time$\downarrow$ & Wins$\uparrow$ & Solved$\uparrow$ \\
\midrule
\multirow{8}{*}{\rotatebox[origin=c]{90}{TSP}}
& MTZ{\footnotesize~\shortcite{miller1960integer}}      & 5.5116 & 0/100 & 79.9319 & 0/100 & 517.6449 & 0/100 & 8/100 \\
& SCF{\footnotesize~\shortcite{gavish1978travelling}}    & 0.8488 & 12/100 & 5.6820 & 15/100 & 129.9687 & 16/100 & 95/100 \\
& MCF-RLT{\footnotesize~\shortcite{balma2018strong}}   & 3.8573 & 0/100 & 56.2487 & 0/100 & 626.8963 & 0/100 & 1/100 \\
\cmidrule{2-9}
& ORLM{\footnotesize~\shortcite{huang2025orlm}}   & 5.5235 & 0/100 & 80.0451 & 0/100 & 517.7556 & 0/100 & 8/100 \\
& StepORLM{\footnotesize~\shortcite{zhou2026steporlm}}   & 4.4893 & 0/100 & 66.7571 & 0/100 & 479.0096 & 0/100 & 13/100 \\
& EvoCut{\footnotesize~\shortcite{yazdani2025evocut}}    & 3.7323 & 0/100 & 27.5892 & 0/100 & 258.2871 & 7/100 & 41/100 \\
\cmidrule{2-9}
& FormuEvo-G    & 0.7488 & 16/100 & 5.3816 & 12/100 & 129.7755 & 15/100 & \textbf{96/100} \\
& FormuEvo (Ours)    & \textbf{0.4517} & \textbf{72/100} & \textbf{3.0088} & \textbf{73/100} & \textbf{47.8586} & \textbf{62/100} & \textbf{96/100} \\
\midrule
\midrule
\multirow{7}{*}{\rotatebox[origin=c]{90}{JSSP}}
& Disj.{\footnotesize~\shortcite{manne1960job}}         & 1.5502 & 12/100 & 9.5882 & 11/100 & 82.3530 & 10/100 & \textbf{97/100} \\
& Enh. Disj.{\footnotesize~\shortcite{ku2016mixed}}        & 2.5150 & 3/100 & 14.1189 & 2/100 & 117.2248 & 3/100 & 93/100 \\
\cmidrule{2-9}
& ORLM{\footnotesize~\shortcite{huang2025orlm}}  & 601.0423 & 0/100 & 600.7678 & 0/100 & 601.7053 & 0/100 & 0/100 \\
& StepORLM{\footnotesize~\shortcite{zhou2026steporlm}}  & 1.6431 & 11/100 & 8.9512 & 9/100 & 77.5248 & 14/100 & 96/100 \\
& EvoCut{\footnotesize~\shortcite{yazdani2025evocut}}   & 1.2713 & 27/100 & 8.0894 & 26/100 & 72.8686 & 20/100 & 96/100 \\
\cmidrule{2-9}
& FormuEvo-G    & 1.5798 & 9/100 & 8.5912 & 14/100 & 77.1278 & 21/100 & \textbf{97/100} \\
& FormuEvo & \textbf{1.1779} & \textbf{38/100} & \textbf{6.7091} & \textbf{38/100} & \textbf{67.7242} & \textbf{32/100} & \textbf{97/100} \\
\bottomrule
\end{tabular}
}
\end{table*}

\begin{table*}[!t]
\centering
\caption{Performance based on SCIP solver.}\label{tab:scip}
\setlength{\tabcolsep}{4pt}
\resizebox{0.9\linewidth}{!}{
\begin{tabular}{ll|cc|cc|ccc}
\toprule
&& \multicolumn{2}{c|}{Easy} & \multicolumn{2}{c|}{Medium} & \multicolumn{3}{c}{Hard} \\
\midrule
& Formulation & Time$\downarrow$ & Wins$\uparrow$ & Time$\downarrow$ & Wins$\uparrow$ & Time$\downarrow$ & Wins$\uparrow$\textsuperscript{*} & Solved$\uparrow$ \\
\midrule
\multirow{8}{*}{\rotatebox[origin=c]{90}{TSP}}
& MTZ{\footnotesize~\shortcite{miller1960integer}}      & 4.7167 & 0/100 & 62.3060 & 1/100 & 563.1162 & 12/100 & 11/100 \\
& SCF{\footnotesize~\shortcite{gavish1978travelling}}    & 4.4918 & 1/100 & 44.2718 & 4/100 & 600.0065 & 0/100 & 0/100 \\
& MCF-RLT{\footnotesize~\shortcite{balma2018strong}}   & 67.1733 & 0/100 & 600.0136 & 0/100 & 600.2085 & 0/100 & 0/100 \\
\cmidrule{2-9}
& ORLM{\footnotesize~\shortcite{huang2025orlm}}   & 4.7421 & 2/100 & 62.0028 & 3/100 & 563.5875 & 12/100 & 11/100 \\
& StepORLM{\footnotesize~\shortcite{zhou2026steporlm}}   & 5.4227 & 3/100 & 84.3814 & 2/100 & 591.2279 & 13/100 & 3/100 \\
& EvoCut{\footnotesize~\shortcite{yazdani2025evocut}}    & 3.6788 & 5/100 & 42.7763 & 10/100 & 555.9330 & 15/100 & 6/100 \\
\cmidrule{2-9}
& FormuEvo-G    & 3.4371 & 11/100 & 35.1599 & 13/100 & 523.7598 & 19/100 & 9/100 \\
& FormuEvo   & \textbf{2.8490} & \textbf{78/100} & \textbf{31.8396} & \textbf{67/100} & \textbf{498.6678} & \textbf{29/100} & \textbf{38/100} \\
\midrule
\midrule
\multirow{7}{*}{\rotatebox[origin=c]{90}{JSSP}}
& Disj.{\footnotesize~\shortcite{manne1960job}}         & 13.0998 & 1/100 & 84.2068 & 1/100 & 491.3035 & 2/100 & 27/100 \\
& Enh. Disj.{\footnotesize~\shortcite{ku2016mixed}}        & 14.3031 & 2/100 & 67.3851 & 1/100 & 405.8829 & 1/100 & 39/100 \\
\cmidrule{2-9}
& ORLM{\footnotesize~\shortcite{huang2025orlm}}  & 600.0982 & 0/100 & 600.0586 & 0/100 & 600.0538 & 0/100 & 0/100 \\
& StepORLM{\footnotesize~\shortcite{zhou2026steporlm}}  & 14.874 & 2/100 & 98.2472 & 1/100 & 486.7095 & 0/100 & 28/100 \\
& EvoCut{\footnotesize~\shortcite{yazdani2025evocut}}   & 7.4703 & 19/100 & 58.4963 & 5/100 & 460.8576 & 3/100 & 30/100 \\
\cmidrule{2-9}
& FormuEvo-G  & 6.5541 & 17/100 & 36.4157 & 14/100 & 341.2865 & 2/100 & 47/100 \\
& FormuEvo & \textbf{4.1672} & \textbf{59/100} & \textbf{17.7541} & \textbf{78/100} & \textbf{114.5116} & \textbf{92/100} & \textbf{94/100} \\
\bottomrule
\end{tabular}
}
\begin{minipage}{0.9\linewidth}
\fontsize{8}{8}\selectfont \textsuperscript{*}For most TSP Hard instances, SCIP fails to solve to optimality within the time limit, making the ``Wins'' metric less indicative.
\end{minipage}
\end{table*}

\subsection{Robustness Across Different Solvers} \label{app:robustness}

To evaluate the robustness and adaptability of FormuEvo across downstream MIP solvers, we replace Gurobi in our main experiments with two alternative solvers: COPT~\cite{ge2022cardinal} and SCIP~\cite{SCIPOptSuite10}, and rerun the evolution process under each solver. The results are reported in Tables~\ref{tab:copt} and \ref{tab:scip}. Here, FormuEvo denotes the formulations evolved directly under the corresponding solver, while FormuEvo-G denotes the formulations evolved under Gurobi (i.e., the one obtained in our main experiments) and then directly adapted to COPT and SCIP, respectively.

\begin{figure*}[!t]
\centering
\includegraphics[width=\linewidth]{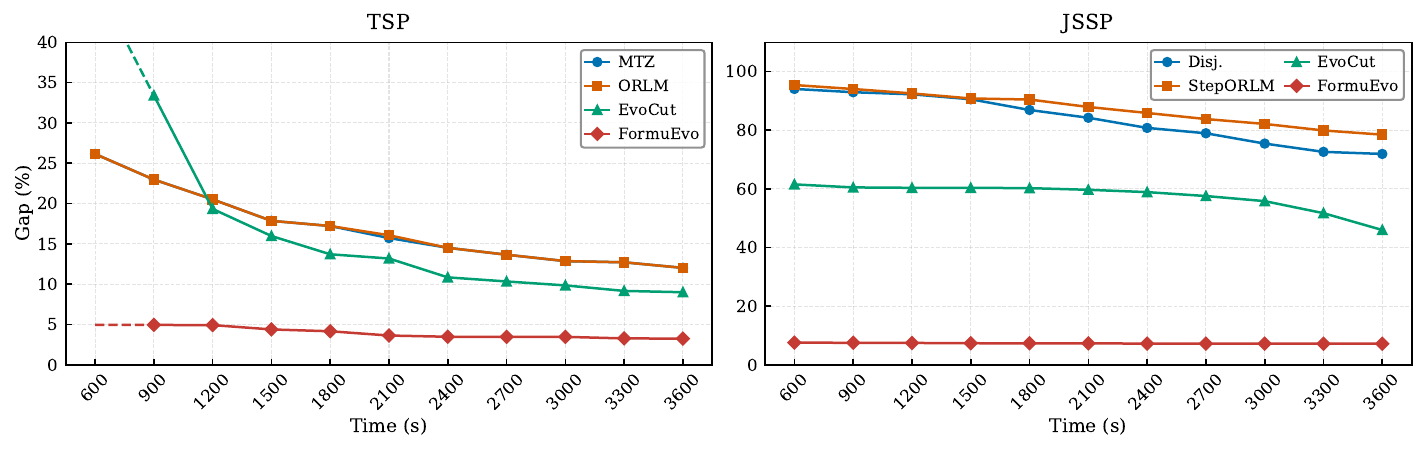}
\caption{Performance comparisons on public benchmark instances. Left: Solomon benchmark of TSP~\cite{solomon1987algorithms}. Right: Taillard benchmark of JSSP~\cite{taillard1993benchmarks}.}
\label{fig:benchmark}
\end{figure*}

Overall, FormuEvo consistently achieves significant improvements on different downstream solvers, demonstrating that the proposed evolutionary framework is not restricted to a specific optimization backend. Meanwhile, we observe substantial variation in the relative performance of different formulations across solvers (e.g., MTZ and MCF-RLT, FormuEvo and FormuEvo-G). This suggests that formulation quality is inherently solver-dependent, and the most effective formulation for one solver may not remain optimal under another optimization backend, due to the algorithmic differences among modern MIP solvers. These results highlight a key advantage of FormuEvo over static formulation generation, and also demonstrate the necessity of our solver-aware optimization that adaptively exploits solver-informed feedback to discover formulations that better align with the algorithmic internals of the downstream solver.

\subsection{Generalization to Public Benchmarks} \label{app:generalization}

We further evaluate the scalability and generalization performance of FormuEvo on large-scale public benchmark instances that are more challenging than the instances used during evolution. For TSP, we test on the 200-city instances from the Solomon benchmark suite~\cite{solomon1987algorithms}. For JSSP, we test on the most challenging $100 \times 20$ instances from the Taillard benchmark~\cite{taillard1993benchmarks}. These public benchmarks, which are more difficult than the Hard instances used in the main experiments, are recognized as challenging large-scale optimization instances for solver evaluation.

Since these instances are generally intractable to solve to optimality within practical time limits, we report the primal-dual gap curves over solving time, where smaller gaps indicate better optimization performance (Fig.~\ref{fig:benchmark}). The results demonstrate that although FormuEvo is evolved using only small instances, the formulations discovered by FormuEvo generalize effectively to benchmark instances with substantially larger scales and different distributions. In particular, the tighter formulations produced by FormuEvo enable the solver to identify high-quality feasible solutions more quickly, thereby reducing the primal-dual gap at a significantly faster rate than baselines.

\subsection{Analysis of Evolutionary Search Hyperparameters} \label{app:hyperparameters}

We conduct a sensitivity analysis of evolutionary search hyperparameters on TSP, including mutation rate $\rho$ and memory usage rate $\gamma$. As shown in Table~\ref{tab:evolutionary_hyperparameters}, FormuEvo
achieves the best overall performance with
$\rho=0.3$ and $\gamma=0.7$, which is selected as our default setting. Notably, a large memory usage rate (e.g., $\gamma=1.0$) does not always lead to better performance, suggesting that occasional memory-free generation helps maintain search diversity and prevents excessive reliance on previously explored patterns.

\begin{table}[!h]
\centering
\caption{Analysis of evolutionary hyperparameters.}
\label{tab:evolutionary_hyperparameters}
\setlength{\tabcolsep}{4pt}
\resizebox{0.7\linewidth}{!}{
\begin{tabular}{lc|c|c|c}
\toprule
\multicolumn{2}{c|}{Param.} & Easy & Medium & Hard \\
\midrule
\multirow{5}{*}{$\rho$}
& 0.0 & 0.1579 & 0.6128 & 4.1886 \\
& 0.3 & \textbf{0.1410} & \textbf{0.5640} & 3.9469 \\
& 0.5 & 0.1903 & 0.6451 & 4.3993 \\
& 0.7 & 0.1862 & 0.6472 & 4.4488 \\
& 1.0 & 0.1456 & 0.5719 & \textbf{3.9436} \\
\midrule
\multirow{5}{*}{$\gamma$}
& 0.0 & 0.1776 & 0.6102 & 4.6649 \\
& 0.3 & 0.1770 & 0.5791 & 4.4793 \\
& 0.5 & 0.1445 & 0.5809 & 4.4923 \\
& 0.7 & \textbf{0.1410} & \textbf{0.5640} & \textbf{3.9469} \\
& 1.0 & 0.1780 & 0.5987 & 4.3957 \\
\bottomrule
\end{tabular}
}
\end{table}

\subsection{Discussion with Existing LLM-based Modeling Approaches} \label{app:discussion}

\begin{figure*}[!t]
\centering
\includegraphics[width=0.85\linewidth]{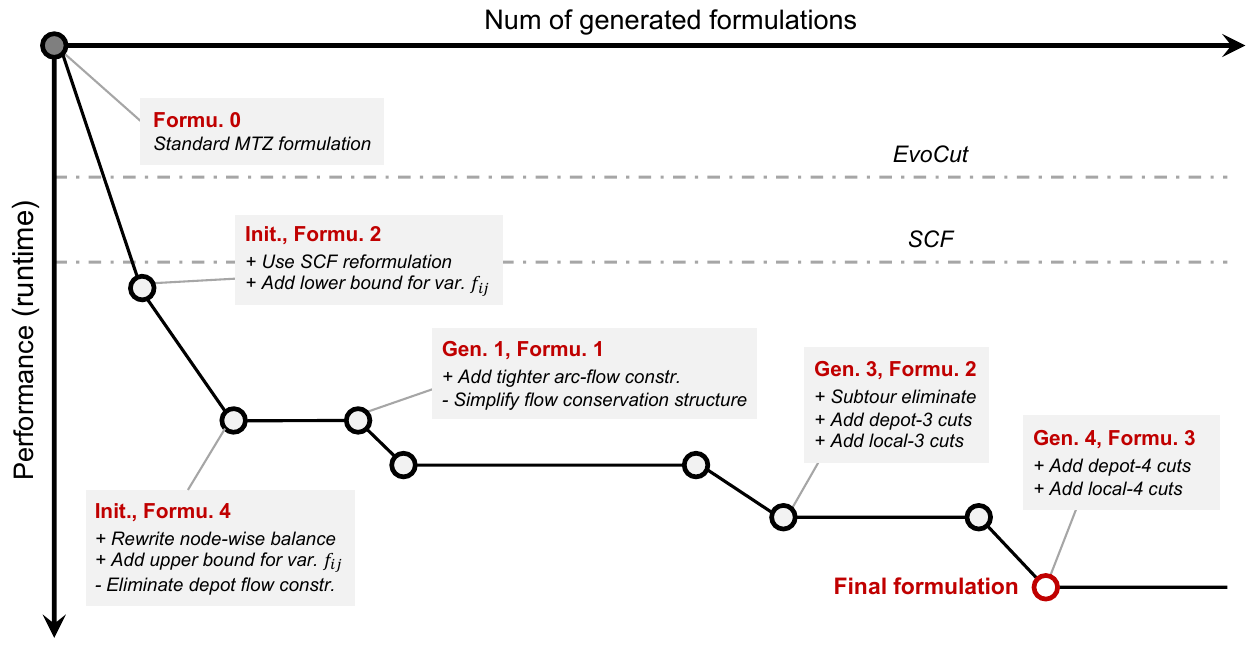}
\caption{An illustration of the evolutionary process of FormuEvo (on TSP).}
\label{fig:evolution}
\end{figure*}

Finally, we provide a discussion about the comparison between FormuEvo and existing LLM-based modeling approaches (e.g., ORLM, StepORLM).

\paragraph{Optimization capability and continual improvement.} Existing LLM-based modeling approaches primarily focus on semantic correctness and syntactical executability through supervised and reinforcement fine-tuning. Although their training data cover a broad range of optimization settings, including linear, integer, and non-linear programming tasks (e.g., Fig.~7 of ORLM~\cite{huang2025orlm}), the capability of these models is largely fixed once fine-tuning is completed. Even when multiple samples are generated during inference, the resulting formulations exhibit very limited diversity (as in our main results): correct generations collapse to nearly the same textbook formulations, while more diverse and efficiency-oriented instructions always lead to incorrect or infeasible results or even semantically meaningless outputs.

In contrast, FormuEvo frames formulation design itself as an explicit optimization problem over the symbolic space of MIP formulations. Instead of relying solely on the static capability of pre-trained models, FormuEvo continuously improves candidate formulations through iterative evolutions guided by solver feedback. As shown in our results and Fig.~\ref{fig:evolution}, even with a small evolutionary budget, FormuEvo can discover high-quality MIP formulations that outperform both expert-designed formulations and recent LLM-based approaches. These results suggest that the formulation design benefits more from the evolutionary search and solver-informed optimization than from static formulation generation alone.

\paragraph{Offline and online computational costs.} Existing fine-tuned modeling approaches rely on extensive curated datasets, long training times, and substantial GPU resources. Their online inference additionally incurs per-instance overhead, since formulations are generated independently for each optimization instance, even though one may attempt to recover a problem-level formulation via post-hoc aggregation using another LLM.

In contrast, FormuEvo requires no task-specific fine-tuning and can be implemented entirely via lightweight API-based interaction with general-purpose LLMs in CPU-only environments. In our experiments, the entire evolutionary process typically requires only one to several hours, depending on the complexity of the target problem and the number of evaluation instances, which is substantially lower than the time of data collection and training required for large-scale fine-tuning. Moreover, FormuEvo operates at the problem level rather than the instance level. Once a strong formulation is evolved for a problem family, it can be directly reused on unseen instances with essentially no additional optimization cost, making FormuEvo particularly attractive in scenarios where many instances have shared underlying optimization structures.

\paragraph{Generality and transferability.} Existing fine-tuned modeling approaches are typically specialized to particular modeling languages, prompting formats, and solver interfaces seen during training, making them extremely sensitive to distribution shifts and implementation-specific overfitting. For example, in ORLM and StepORLM, we find that even minor modifications from their prompt templates can lead to substantial degradation or complete failure of the generated formulations. In contrast, FormuEvo operates on top of general-purpose foundation LLMs and treats optimization formulations as symbolic programs that can be flexibly modified and evaluated. Therefore, its optimization capability is not tied to a specific LLM backbone, modeling language, solver backend, or benchmark distribution (Appendix~\ref{app:robustness}~-~\ref{app:hyperparameters}).  The evolution framework and distilled knowledge can naturally extend across different optimization paradigms, modeling interfaces, and solver implementations, suggesting broader applicability and stronger transferability in practical optimization settings.

\section{The Use of LLMs}

LLMs are used as key methodological components for evolutionary search in our framework. We also used LLMs to assist code writing and text polishing, while the core ideas and the manuscript itself were conceived, prepared, and finalized by the authors.

\end{document}